\documentclass[english, 10pt, twocolumn, twoside]{IEEEtran}
\usepackage{threeparttable}
\usepackage{mathrsfs}

\usepackage{cite}

\ifCLASSINFOpdf
\else
\fi
\usepackage{graphics}

\usepackage{subfigure}
\usepackage[cmex10]{amsmath}
\usepackage{amsfonts}
\usepackage{cases}
\usepackage{epsfig}
\usepackage{url}

\usepackage{algorithm}
\usepackage{algorithmic}

\usepackage{array}
\usepackage{amsmath}

\usepackage{multirow}
\usepackage{bigstrut}

\usepackage{bm}
\usepackage{amsmath}
\usepackage{amssymb}
\usepackage{mathrsfs}
\usepackage{color}

\usepackage{booktabs}
\usepackage{multirow}
\usepackage{amsmath}
\usepackage[bottom]{footmisc}
\usepackage{bbm}
\usepackage{hyperref}
\hypersetup{
	colorlinks=true
}

\newcommand{\comment}[1]{\textcolor[rgb]{0.0,0.00,0.00}{#1}}

\newcommand {\bt}[1]{\bf{#1.}\normalfont}

\begin{document}

%
\title{RViDeformer: Efficient Raw Video Denoising Transformer with a Larger Benchmark Dataset}
%
%
%

\author{Huanjing Yue, \textit{ Senior Member, IEEE}, Cong Cao, Lei Liao, Jingyu Yang, \textit{Senior Member, IEEE}
\thanks{This work was supported in part by the National Natural Science Foundation of China under Grant 62472308 and Grant 62231018.
H. Yue, C. Cao, L. Liao, and J. Yang (corresponding author) are with the School of Electrical and Information Engineering, Tianjin University, Tianjin, China, 300072.}}

%
%

\markboth{IEEE TRANSACTIONS ON CIRCUITS AND SYSTEMS FOR VIDEO TECHNOLOGY}%
{Yue \MakeLowercase{\textit{et al.}}: RViDeformer: Efficient Raw Video Denoising with a Larger Benchmark Dataset}
%



\maketitle

\begin{abstract}
In recent years, raw video denoising has garnered increased attention due to the consistency with the imaging process and well-studied noise modeling in the raw domain. However, two problems still hinder the denoising performance. Firstly, there is no large dataset with realistic motions for supervised raw video denoising, as capturing noisy and clean frames for real dynamic scenes is difficult. To address this, we propose recapturing existing high-resolution videos displayed on a 4K screen with high-low ISO settings to construct noisy-clean paired frames. In this way, we construct a video denoising dataset (named as ReCRVD) with 120 groups of noisy-clean videos, whose ISO values ranging from 1600 to 25600. Secondly, while non-local temporal-spatial attention is beneficial for denoising, it often leads to heavy computation costs. We propose an efficient raw video denoising transformer network (RViDeformer) that explores both short and long-distance correlations. Specifically, we propose multi-branch spatial and temporal attention modules, which explore the patch correlations from local window, local low-resolution window, global downsampled window, and neighbor-involved window, and then they are fused together. We employ reparameterization to reduce computation costs. Our network is trained in both supervised and unsupervised manners, achieving the best performance compared with state-of-the-art methods. Additionally, the model trained with our proposed dataset (ReCRVD) outperforms the model trained with previous benchmark dataset (CRVD) when evaluated on the real-world outdoor noisy videos. Our code and dataset are available at \href{https://github.com/cao-cong/RViDeformer}{https://github.com/cao-cong/RViDeformer}.
\end{abstract}


%
\IEEEpeerreviewmaketitle

\section{Introduction}
\label{sec:introduction}
%
%
%
%
Noise is inherent to every imaging sensor, which not only degrades the visual quality but also affects the following understanding and analysis tasks. Generally, video denoising can achieve better results than single-frame based denoising due to the temporal correlations between neighboring frames. In addition, the noise distribution in raw domain (the directly output of the sensor) is much simpler than that in sRGB domain due to the nonlinear image signal processor (ISP).  Therefore, raw video denoising is attractive for improving the video imaging quality.


However, there is still some challenges in raw video denoising. The first problem is how to build a large scale video denoising dataset.
Building a benchmark dataset with paired noisy-clean raw videos is important for supervised training and evaluating both supervised and unsupervised raw video denoising methods. However, capturing paired noisy-clean raw videos is challenging. For image denoising, the ground truth is usually captured with long exposure time or via averaging a lot of noisy images to get a clean image. However, this will lead to motion blur or large displacements when applied to video ground truth capturing.  
The work in \cite{yue2020supervised} tried to solve the problem  by capturing with stop-and-motion manner, which captured each noisy-clean pair with a static scene, and concatenated frames with temporal order to generate paired noisy-clean raw videos, building the CRVD dataset. 
However, the motions in CRVD are unnatural and the scene number is limited (11 scenes). 
The work in \cite{kong2020comprehensive} moved cameras rather than moving the objects to create translation motions, and they only captured sRGB video pairs. Neither moving toys nor moving cameras can simulate the various motions in real world. Training with the two datasets limits the model's ability in dealing with real-world noisy videos. Based on the above observations, we propose to recapture the existing high-resolution videos displayed on a 4K screen to construct our dataset. Specifically, we recapture the screen content with high and low ISO settings to construct noisy-clean paired frames, and concatenate them with temporal order to construct paired videos. In this way, our recaptured raw video denoising (ReCRVD) dataset contains various real motions and a large amount of scenes.

When this paper was written, we became aware of a very recent work \cite{fu2022low} which also utilizes the similar way to capture noisy-clean video pairs displayed on the screen. But \cite{fu2022low} focuses on low-light raw video denoising and actually captures low-normal light video pairs. The ISO setting is fixed to 100 and the noisy frames are constructed by linearly scaling the low-light  frames to match the brightness of normal-light frames. Therefore, the analog gain of the camera is small and the noise in their dataset mainly depends on the digital gain of the camera. In contrast, we directly capture noisy-clean frames with almost the same brightness and the ISO settings cover a large range (i.e., the analog gain spanning a large range), which is more consistent with general video capturing. In this way, our work focuses on general raw video denoising other than low light video enhancement.



Besides the dataset, we also explore raw video denoising methods. Recently, transformer-based video denoising methods \cite{song2022tempformer,liang2024vrt,liang2022recurrent,fu2022low} have shown promising performance. These networks are based on shift-window self-attention from Swin Transformer \cite{liu2021swin,liang2021swinir}, which divides input images into non-overlapping windows and the attention is calculated inside the window. The information of neighboring windows is exchanged through window shift operation. Since it forces the self-attention in a window, the long-range spatial and temporal information cannot be utilized for video denoising. Based on the above observations, we propose Multi-branch Spatial Self-Attention (MSSA) and Multi-branch Temporal Self-Attention (MTSA) modules. Besides the plain shift-window self-attention (SWSA) branch, we propose Low-Resolution-Window Self-Attention (LWSA), Global-Window Self-Attention (GWSA) and Neighbour-Window Self-Attention (NWSA) branch for spatial reconstruction, where the queries are the same as that in plain SWSA, while the keys and values of the transformer are constructed by tokens in the low-resolution window, downsampled global window, and neighbor involved window, respectively. The long-range temporal correlations are also explored in a similar way. The additional self-attention branches enable the network to utilize the information from long-range similar patches in spatial and temporal dimensions. The information from different self-attention branches are balanced and fused in the proposed multi-branch architecture. Besides the improvement of self-attention mechanism, we combine linear layer with convolution in the bottom of each block to increase the receptive field. We further utilize reparameterization to reduce the computation cost. In this way, we propose an efficient transformer network  for raw video denoising (RViDeformer).

We would like to point out that the process of generating noisy-clean raw video pairs is labor-intensive. If we lack access to noisy-clean pairs, can our network still produce satisfactory results? Leveraging advances in unsupervised loss functions for image denoising \cite{huang2021neighbor2neighbor}, we find that applying unsupervised loss directly to our proposed network yields satisfactory performance. This highlights the efficacy of our network for both supervised and unsupervised denoising. Additionally, our unsupervised method exhibits superior generalization performance on real-world outdoor noisy videos. Our contributions are summarized as follows


\begin{itemize}
\item{First, we construct a large scale raw video denoising dataset (named as ReCRVD, including 120 scenes) by recapturing the videos displayed on the screen. To make the noisy-clean pairs to be pixel-aligned and approximate outdoor capturing, we propose intensity correction, spatial alignment, and color correction for post-processing. The model trained on ReCRVD generalizes better to real-world outdoor noisy videos compared with that trained on the widely used CRVD.\comment{To our knowledge, our dataset is the first recaptured raw video dataset that targets at general raw video denoising with various ISO settings.}}


\item{\comment{Second, we propose an efficient raw video denoising transformer network (RViDeformer) by designing different windows in transformer. In order to explore long-range correlations with modest computation costs, we propose neighbor window, and global down-sampled window for attention calculation. To utilize multiscale information inside the transformer block, we further propose low-resolution window. They construct our multi-branch spatial and temporal attention modules to capture the short and long distance correlations efficiently.}}


\item{Third, our network is trained in both supervised and unsupervised manners, and they achieve the best performance on our proposed ReCRVD dataset and CRVD indoor dataset with the smallest computation cost when compared with state-of-the-art denoising methods. In addition, our method has better generalization performance on the real-world outdoor noisy videos.}
\end{itemize}

\section{Related Works}

\subsection{Supervised Video Denoising}


Different from image denoising \cite{dong2018denoising,guo2019toward,chen2019real,zhang2020residual,jiang2022deep,pan2022real,zhou2023deep,lu2023virtual}, video denoising \cite{buades2019cfa,sun2023deep} utilize both spatial and temporal correlations in the noisy frames. Traditional video denoising method VBM4D \cite{matteo2011video} exploits the mutual similarity between 3D spatio-temporal volumes and filters the volumes according to the sparse representation. For CNN-based methods, \cite{chen2016deep} first applies CNN to video denoising and the information from previous frame is utilized based on a recurrent network. Xue \textit{et al.} \cite{xue2019video} proposed a task-oriented flow (TOFlow) to align neighbour frames to the current frame. Yue \textit{et al.} \cite{yue2020supervised} proposed to utilize temporal fusion, spatial fusion, and non-local attention to fully explore the correlations between neighboring frames. Besides, efficient video denoising is attracting more and more attention, such as the \comment{binarized low-light raw video denoising \cite{zhang2024binarized}}, efficient multi-stage video denoising \cite{maggioni2021efficient}, bidirection recurrent network with look-ahead recurrent module \cite{li2022unidirectional}, and FastDVDnet \cite{tassano2020fastdvdnet}, which is constructed by two UNet denoisers. \comment{Inspired by the two denoiser structure, \cite{qi2022real} and \cite{li2023simple} further utilize temporal shift and grouped spatial-temporal shift for temporal fusion, respectively.}


Recently, transformer-based video denoising methods \cite{song2022tempformer,liang2024vrt,liang2022recurrent,fu2022low} have shown promising performance.
Song \textit{et al.} \cite{song2022tempformer} proposed joint Spatio-Temporal Mixer for each transformer block to aggregate features. Liang \textit{et al.} \cite{liang2024vrt} proposed Temporal Mutual Self-Attention to exploit temporal information. Then, they \cite{liang2022recurrent}  further combine transformer with recurrent network
 and flow-guided deformable attention. Recently, Fu \textit{et al.} \cite{fu2022low} proposed a low-light raw video denoising network based on 3D (Shifted) Window-based Multi-head Self-attention. All the denoising transformers are based on Swin Transformer \cite{liu2021swin,liang2021swinir}, which calculates attention inside a local window and information between different windows exchanges through window shift operation. The local-window based attention restricts the long-distance correlation utilization. In this work, we propose global-window attention and neighbour-window attention to enable exploiting information from global and neighbour context and utilize multi-branch architecture to comprehensively utilize different attention mechanisms.


\subsection{Unsupervised Video Denoising}

Since supervised denoising relies on expensive paired noisy-clean data collection, unsupervised denoising methods have been proposed to alleviate this problem.
Representative unsupervised image denoising methods include Noise2Noise \cite{lehtinen2018noise2noise}, Noise2Void \cite{krull2019noise2void}, R2R \cite{pang2021recorrupted}, and NBR2NBR \cite{huang2021neighbor2neighbor}. These methods either utilize another noisy image \cite{lehtinen2018noise2noise} or regenerated noisy image \cite{pang2021recorrupted} \cite{huang2021neighbor2neighbor} as labels, or utilize blind-spot strategy, which predicts the center pixel from its context noisy pixels \cite{krull2019noise2void}.


For unsupervised video denoising, noisy-noisy pairs can be constructed by warping neighboring frames. F2F \cite{ehret2019model} utilizes optical flow to warp the neighbor noisy frame to serve as the label of target frame, and utilizes image denoising network DnCNN \cite{zhang2017beyond} for video denoising. On the basis of F2F, MF2F \cite{dewil2021self} designs a multi-input network, which selects different neighbor frames for network inputs and labels, respectively. However, since the warping process will introduce errors, the performances of F2F and MF2F are limited. UDVD \cite{sheth2021unsupervised} utilizes half-plane convolution to construct blind-spot video denoising network. \comment{TAP \cite{fu2024temporal} integrates tunable temporal modules into a pre-trained image denoiser and proposes a progressive fine-tuning strategy to refine the temporal module using pseudo labels.} Different from them, we utilize our proposed network for unsupervised denoising, and utilize NBR2NBR \cite{huang2021neighbor2neighbor} strategy to get two sub-frames to construct the noisy-noisy pairs.


\textcolor[rgb]{0.0,0.00,0.00}{Besides unsupervised video denoising, zero-shot video denoising \cite{cao2024zero} has also been proposed to avoid the expensive collection of paired noisy-clean data. The work in \cite{cao2024zero} adapts the diffusion-based image denoising model to temporally consistent video denoising without additional training. However, its performance may be severely degraded when encountering new noisy video due to lack of training. In contrast, our proposed method can still perform well through unsupervised fine-tuning on the new noisy video data.}

\subsection{Image and Video Processing with Raw Data}
During image capturing, the raw data collected by sensors usually goes through a complex ISP module (including demosaicing, white balance, tone mapping \textit{etc} ) to generate the final sRGB image. Without complex nonlinear transform and quantization, the noise distribution in raw domain is much simpler and the raw image has wider bit depth (12/14 bits per pixel). Therefore, many image reconstruction works turn to raw domain processing and have achieved better performance, such as image (video) super-resolution \cite{xu2019towards,zhang2019zoom,yue2022real}, joint restoration and enhancement \cite{ratnasingam2019deep,schwartz2018deepisp,liang2019cameranet,ignatov2020aim}, image deblurring \cite{liang2020raw}, image demoir\'eing \cite{yue2022recaptured}.



For denoising, many raw image denoising methods have been proposed \cite{gharbi2016deep,chen2019learning,liu2019learning,abdelhamed2019ntire,abdelhamed2020ntire}, and several raw image denoising datasets \cite{anaya2014renoir,abdelhamed2018high,plotz2017benchmarking,chen2019learning} were constructed. Since sRGB images are more common in our daily life, Brooks \textit{et al.} \cite{brooks2018unprocessing} proposed a simple inverse ISP method to unprocess sRGB images back to the raw domain, which is helpful to generate more training data to improve raw image denoising performance. Similarly, Zamir \textit{et al.}  \cite{zamir2020cycleisp} proposed to use a CNN to learn inverse ISP to better synthesize raw noisy-clean pairs. Besides directly using noise synthesis for data augmentation, Liu \textit{et al.} \cite{liu2019learning} proposed Bayer pattern unification and Bayer preserving augmentation method and achieved the winner of NTIRE 2019 Real Image Denoising Challenge. Considering the huge computing cost of previous denoising networks, Wang \textit{et al.} \cite{wang2020practical} proposed a lightweight model which is designed for raw image denoising on mobile devices.

However, there are only a few works dealing with raw video denoising due to the unavailable of dynamic video sequence pairs.
Chen \textit{et al.} \cite{chen2019seeing} proposed to transform low-light noisy raw frames to the normal-light sRGB ones and their dataset is constructed by static sequences. RViDeNet \cite{yue2020supervised} proposed to pack the noisy Bayer videos into four branches and perform denoising separately and then combine them together. In this work, we construct a larger raw video denoising dataset and propose an efficient transformer-based raw video denoising network.



\subsection{Real-World Image and Video Denoising Datasets}

In order to benchmark realistic noise removal, many paired noisy-clean image denoising datasets have been proposed. The clean image is usually captured with long exposure (RENOIR \cite{anaya2014renoir}, DND \cite{plotz2017benchmarking} and SMID \cite{chen2019seeing} datasets) or by averaging many noisy shots \cite{nam2016holistic,yue2019high,xu2018real,abdelhamed2018high}. The captured real noisy images in these datasets are saved either in sRGB format \cite{nam2016holistic,yue2019high,xu2018real} or raw format \cite{anaya2014renoir, plotz2017benchmarking, abdelhamed2018high,chen2019learning}, whose sRGB images are generated by simple ISP algorithms.



Real-world video denoising datasets are relatively scarce compared to image denoising datasets since capturing clean frames for dynamic scenes in real-time is challenging.
The work in \cite{chen2019seeing,xu2022pvdd} solved this problem by constructing static videos with no dynamic objects, whose ground truth can be directly generated by frame averaging. In our previous work \cite{yue2020supervised}, we constructed a raw video denoising dataset (CRVD) by introducing the stop-and-motion capturing method, which captures each static scene many times to generate a clean frame, and repeats this process after moving the objects. Similarly, IOCV dataset \cite{kong2020comprehensive} is constructed by moving cameras automatically rather than moving the objects, but the frames are saved in sRGB domain. 
However, both datasets have simple motions that differ from real-world motions. Another strategy is to utilize a beam splitter in a co-axis optical system \cite{jiang2019learning} to capture realistic motions in the wild. However, the photons are divided in half by the beam splitter, making the ground truth frames affected by noise and thus limiting the quality of the dataset.
In this work, we address the issue by recapturing existing videos displayed on a screen using high-low ISO settings to create noisy-clean pairs frame by frame. This capturing approach resulted in a larger dataset with diverse motions, which will facilitate future research on raw video denoising. While \cite{fu2022low} also utilizes screen recapturing to construct the dataset, their dataset is designed for low light video enhancement while our dataset is designed for general video denoising, as described in Sec. \ref{sec:introduction}.

\section{ReCRVD Dataset Construction}

\subsection{Capturing Procedure}



As mentioned in Sec. \ref{sec:introduction}, the main challenge for creating a video denoising dataset is capturing both noisy and clean frames for dynamic scenes concurrently. 
%
%
%
We observe that utilizing screens for paired image capturing \cite{peng2019learned} is a good substitution when the ground truth is difficult to be collected. 
Therefore, we propose capturing noisy-clean pairs by sequentially displaying existing high-resolution video frames on a 4K screen and recapturing the screen content. For each displayed video frame, we randomly select one ISO value from a set of five settings (1600, 3200, 6400, 12800, 25600) and continuously capture ten noisy samples with the selected ISO and short exposure time. Subsequently, we capture one clean frame with low ISO (100) and long exposure time. After capturing all frames for the current video, we group them according to their temporal order to generate the dynamic noisy video and its corresponding clean video. Note that, capturing ten noisy samples for each displayed frame increases the diversity of noise samples in our dataset. We employ a surveillance camera equipped with the IMX385 sensor, identical to the one utilized in the CRVD dataset \cite{yue2020supervised}, for video capturing. The raw image sequences are captured at a rate of 20 frames per second, and the resolution for the Bayer frame is $1920\times1080$. The displayed videos include 100, 16, and 4 high-quality videos from the DAVIS \cite{perazzi2016benchmark}, UVG \cite{mercat2020uvg}, and Adobe240fps \cite{su2017deep} datasets, respectively. The frame rates for the three datasets are 30 fps, 50/120 fps, and 240 fps, respectively. As a result, our dataset includes videos with various frame rates, which enhances its generalization in real-world scenarios.




\begin{figure}[htb]
    \centering
    \includegraphics[width=1.0\linewidth]{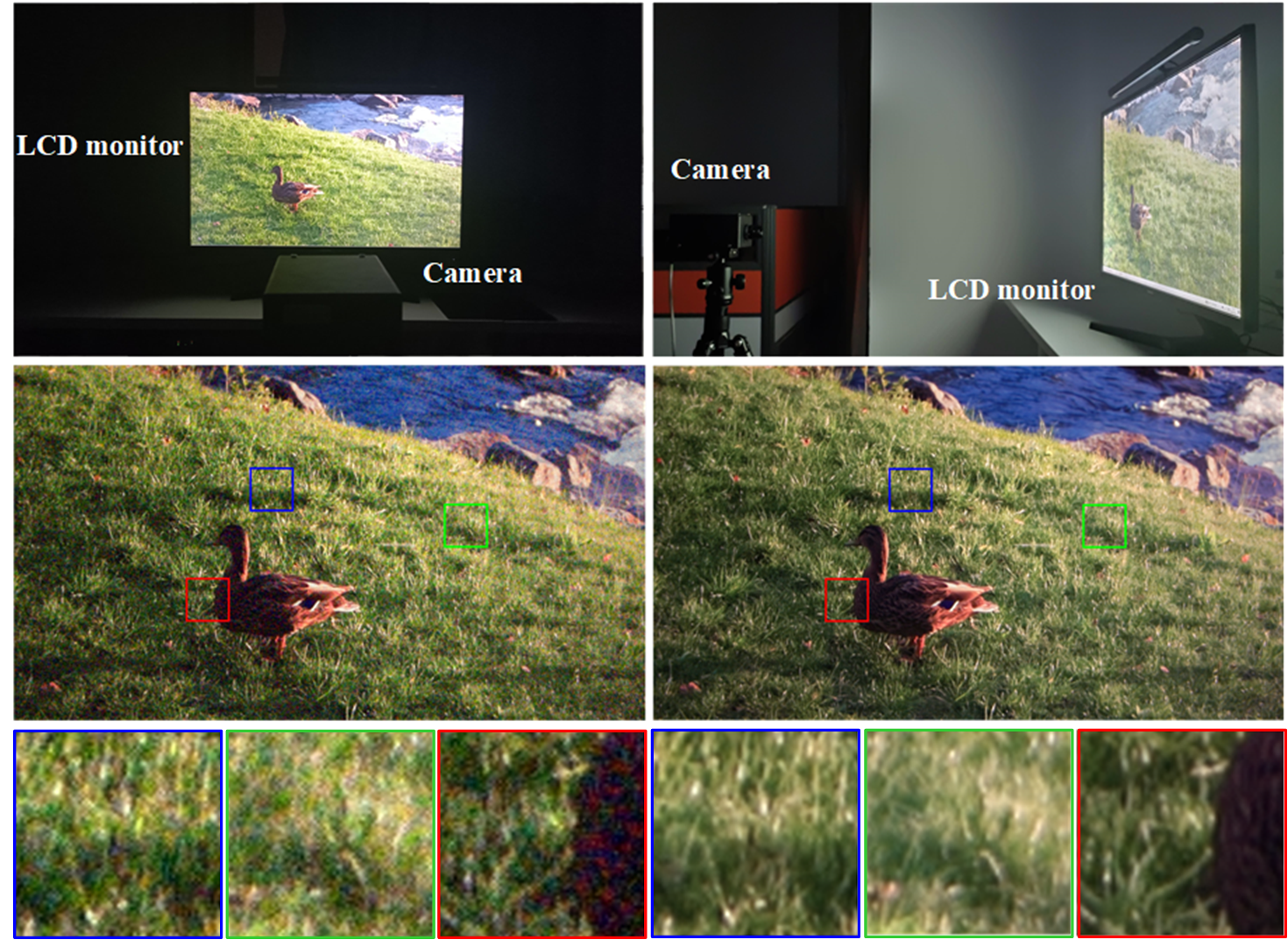}
    \caption{Illustration of our capturing system (top row), the captured noisy-clean pairs (middle row), and the close-up of three regions (bottom row).}
    \label{fig:device}
\end{figure}

It should be noted that recapturing the screen content may introduce unwanted moir\'e patterns due to the aliasing between the grids of the display screen and color filter array (CFA) in the camera sensor. Fortunately, Thongkamwitoon \textit{et al.} \cite{thongkamwitoon2015image} have demonstrated that such patterns can be prevented by adjusting the diaphragm aperture or focal length. In practice, we place the camera and screen as shown in Fig. \ref{fig:device}. We maintain a constant diaphragm aperture and carefully adjust the focal length to prevent moir\'e patterns and lens blur in the recaptured content. Fig. \ref{fig:device} displays a recaptured noisy frame and its corresponding noise-free frame, along with a close-up view of the difference between the noisy and clean frames. It can be observed that the difference is pixel-independent noise without any characteristic color stripes, i.e., moir\'e patterns.

To eliminate the influence of ambient lighting, our dataset is captured in a darkroom. We adjust the brightness of the display screen to ensure that the luminance around the camera is low, approximately 1 lux. We captured 120 pairs of dynamic noisy-clean videos under five different ISO levels ranging from 1600 to 25600, with 24 videos captured for each ISO value. These 120 scenes are divided into a training set (90 scenes) and a testing set (30 scenes). 
Table \ref{Dataset} lists the summary information of the proposed ReCRVD dataset and compares it with the widely used CRVD \comment{indoor} dataset. It can be observed that the ReCRVD dataset is richer than the CRVD indoor dataset. \textcolor{black}{Please refer to our supplementary file for more details about our dataset.} 

We would like to point out that it is difficult to capture different noise levels under arbitrary environmental conditions. For practical, we randomly selecting a set of ISO values to capture different noise levels, which is a common strategy in the construction of denoising datasets \cite{abdelhamed2018high,plotz2017benchmarking,jiang2019learning}. The captured noise is more realistic than simulated noise. Recently, there are some works focusing on simulating noise under different environmental conditions, such as \cite{wei2020physics,zhang2021rethinking,feng2022learnability}. However, they are also trying to learn statistic information from real captured noisy frames. Although we cannot capture under all the environmental conditions in the real-world, our captured dataset is a good representation for a large range of noise levels. In addition, these noisy-clean videos can also help the learning process for noise simulation \cite{wei2020physics,zhang2021rethinking,feng2022learnability}. It would be a good research topic to combine the two strategies, namely capturing noisy-clean pairs and simulating noise under arbitrary conditions, to construct more realistic denoising datasets. 





\begin{table}[t]
\centering
\caption{Comparison of the CRVD \comment {indoor} dataset and the proposed ReCRVD dataset.}
\begin{tabular}{m{3.2cm}m{1.9cm}<{\centering}m{1.0cm}<{\centering}}
\toprule
Methods                               & CRVD \comment{Indoor}     & ReCRVD \\
\hline
Total scenes                          & 11       & 120      \\
Training scenes                       & 6        & 90       \\
Testing scenes                        & 5        & 30       \\
Number of noise levels                & 5        & 5        \\
Number of paired videos               & 55       & 120      \\
Realistic motion                      & $\times$   & $\checkmark$ \\
\bottomrule
\end{tabular}
\label{Dataset}
\end{table}


\subsection{Post Processing}
There often exists brightness difference and spatial misalignment between the captured noisy frame and the corresponding clean frame due to the different ISO settings, exposure times, illuminations, and the minuscule destabilization of the camera.
Therefore, we further apply post-processing to our captured data to correct brightness difference, align the clean frame to the noisy frame, and correct the color cast.

\begin{figure}[htb]
    \centering
    \includegraphics[width=1.0\linewidth]{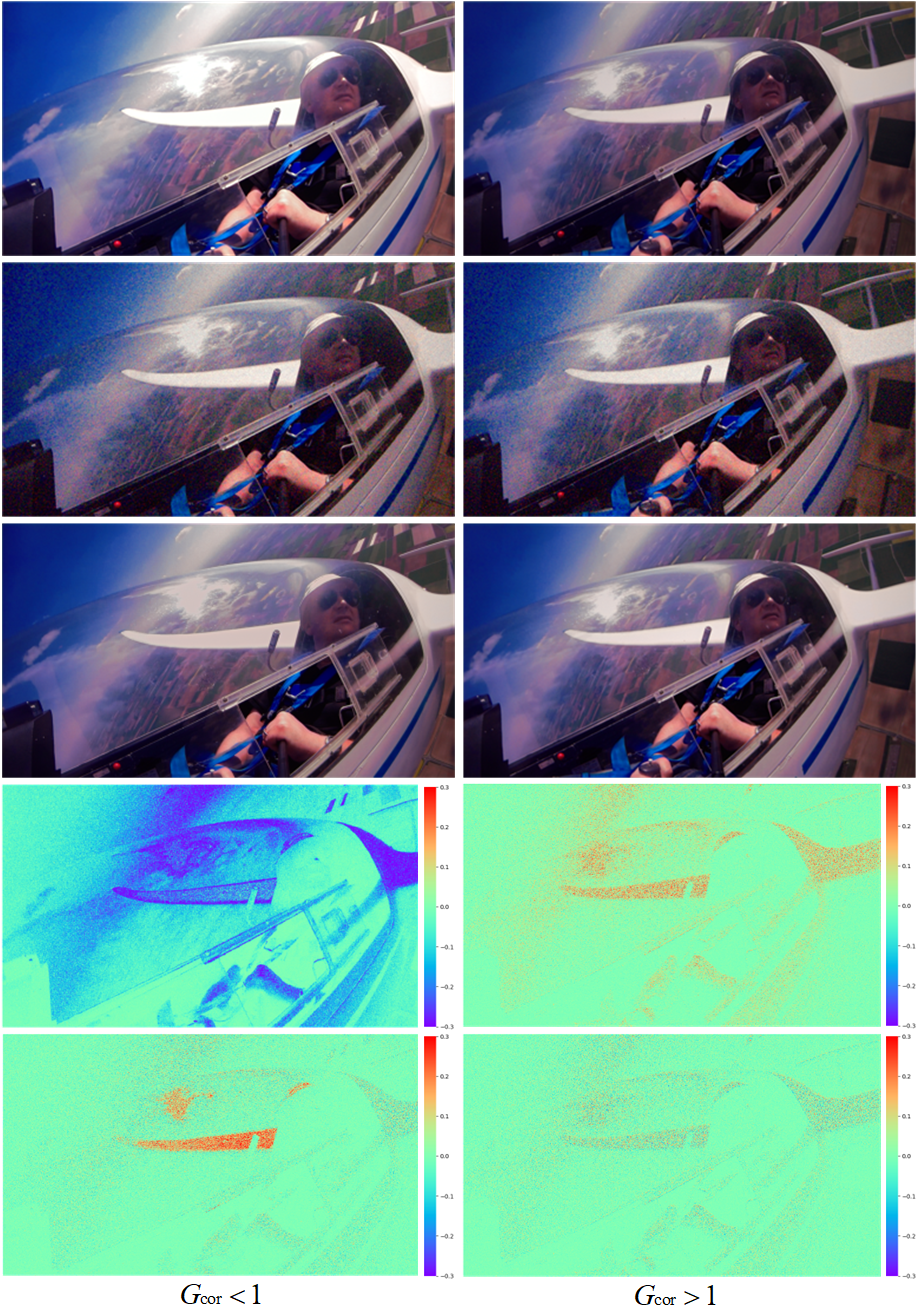}
    \caption{From top to bottom, each row lists the original clean frame, noisy frame, clean frame after intensity correction, the residual brightness channel before and after intensity correction. Left: processing with $G_\text{cor}<1$; Right: processing with $G_\text{cor}>1$ (proposed).}
    \label{fig:intensitycorrection}
\end{figure}

\subsubsection{Linear Intensity Correction}
We capture the displayed videos in a darkroom and keep screen luminance as a constant value so that the brightness of the captured raw frame is only influenced by ISO settings and exposure time, which can be formulated as
\begin{equation}
B = G\times E,
\label{brightness}
\end{equation}
where $B$ is the brightness of the raw frame pixels, $G$ is the ISO gain and $E$ is the exposure time. We denote the brightness, ISO gain, exposure time for raw clean (noisy) image as $B_{\text{c}}$ ($B_{\text{n}}$), $G_{\text{c}}$ ($G_{\text{n}}$), and $E_{\text{c}}$ ($E_{\text{n}}$), respectively. In order to not change the distribution of the noise in noisy frame, we correct the clean raw frame by multiplying it with a brightness compensation coefficient $G_\text{cor}$ to make the clean frame have the same brightness as noisy frame. This process can be formulated as
\begin{equation}
B_{\text{c}}\times G_\text{cor} = B_{\text{n}}.
\label{brightnessequal}
\end{equation}
For a raw image, the pixel value after black level correction is linearly to $B$. In other words, the average brightness of a raw image can be estimated by averaging the pixel values.  Therefore, $G_\text{cor}$ can be derived as follows
\begin{equation}
G_\text{cor} = \frac{B_{\text{n}}}{B_{\text{c}}} = \frac{ \sum_{i,j}(\mathbf{I}_{ij}^{\text{n}}-\text{bl})}{ \sum_{i,j}(\mathbf{I}_{ij}^{\text{c}}-\text{bl})}
\label{intensitycorrection}
\end{equation}
where $\mathbf{I}_{ij}^{\text{n}}$ ($\mathbf{I}_{ij}^{\text{c}}$) denotes the pixel value of the raw noisy (clean) image at coordinates $(i,j)$, and $\text{bl}$ is the black level. Note that, the over-exposed pixel values are clipped to white level and is not linearly to $B$. When $G_\text{cor}$ is smaller than 1, directly applying $G_\text{cor}$ to $\mathbf{I}^\text{c}$  will cause the over-exposed pixel values in $\mathbf{I}^\text{c}$ become smaller than the white level. As shown in the left column of Fig. \ref{fig:intensitycorrection}, the residual B channel (obtained by $\mathbf{I}^\text{n}$-$\mathbf{I}^\text{c}$ (${\hat{\mathbf{I}}^\text{c}}$), where ${\hat{\mathbf{I}}^\text{c}}$ is the intensity corrected version) has large difference at the over-exposed regions, such as the person's hat and the wings of air plane. Therefore, during capturing, we tune the exposure time and screen luminance to make the recaptured clean image a bit darker than the noisy image. In this way, the derived $G_\text{cor}$ is larger than 1. As shown in the right column of Fig. \ref{fig:intensitycorrection}, after intensity correction, ${\hat{\mathbf{I}}^\text{c}}$ has the same brightness as that of $\mathbf{I}^\text{n}$, even in the over-exposed regions.

We would like to point out that the exposure compensation method may not be physically accurate. But this is the most practical strategy, and our experiments (in the supplementary file) also demonstrate the superiority of our dataset.



\subsubsection{Spatial Alignment}
\begin{figure}[htb]
    \centering
    \includegraphics[width=1.0\linewidth]{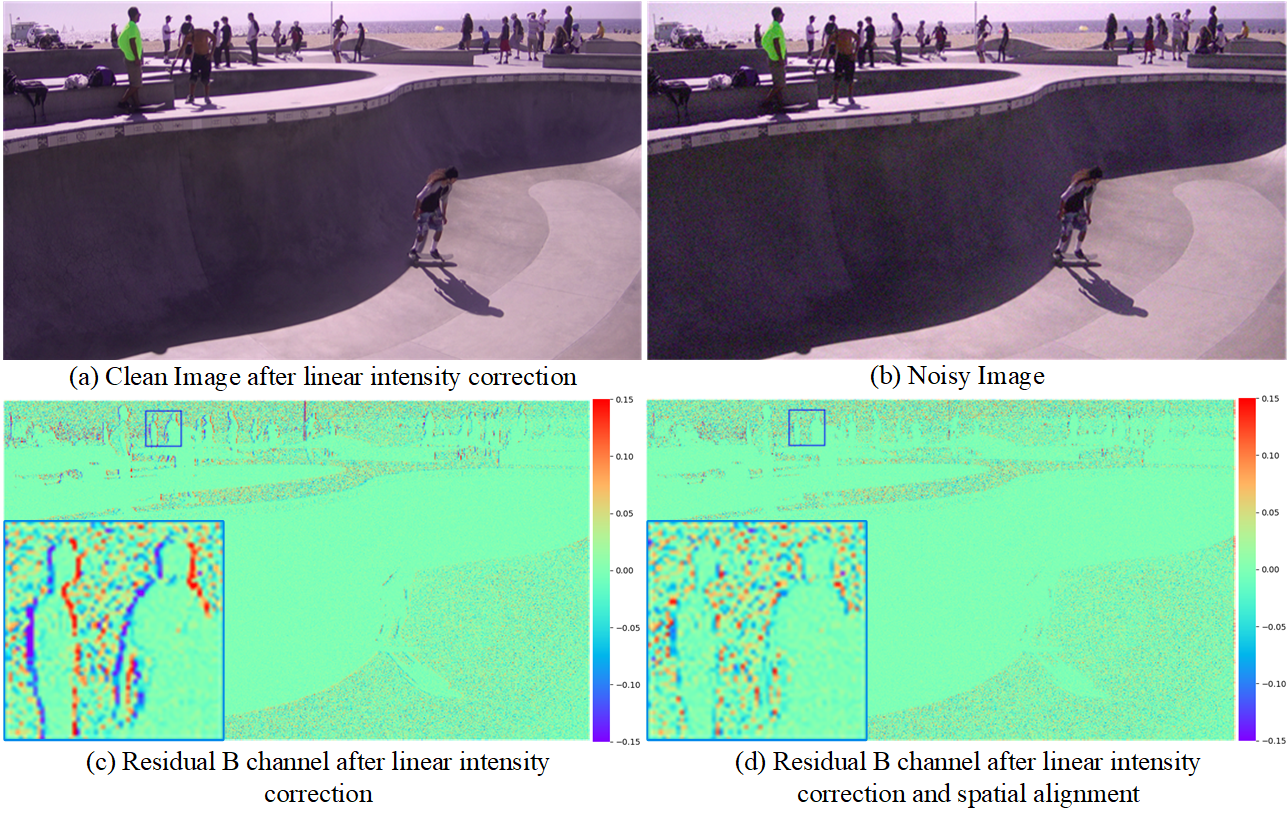}
    \caption{Differences between blue channels of noisy and clean raw frames before and after spatial alignment. Zoom in for better observation.}
    \label{fig:spatialalignment}
\end{figure}

Due to the minuscule destabilization of the camera, there are spatial misalignments between $\mathbf{I}^\text{n}$ and ${\hat{\mathbf{I}}^\text{c}}$. In this work, we utilize DeepFlow \cite{weinzaepfel2013deepflow} to align
${\hat{\mathbf{I}}^\text{c}}$ with $\mathbf{I}^\text{n}$.
Specifically, we utilize DeepFlow to compute the optical flow between Gr channel of ${\hat{\mathbf{I}}^\text{c}}$ and  $\mathbf{I}^\text{n}$, and then apply the same flow to the packed four channels of ${\hat{\mathbf{I}}^\text{c}}$ to perform the warping. In this way, the Bayer pattern of the raw image can be kept. From Fig. \ref{fig:spatialalignment}, it can be observed that there exists sharp object edge in (c) but they disappear in (d). It means that after spatial alignment, raw clean frame is well aligned to raw noisy frame. 
Note that, the spatial reinterpolation with subsampled 4-channel CFA may lead to aliasing problems in high-frequency regions. Fortunately, we checked our dataset and did not find these problems. All warping frames are manually carefully checked  and the ones with alignment errors are removed from our dataset.


\subsubsection{Color Correction}


Due to the blue light of the electronic monitor, the images captured by the camera may have color cast, which will make the data set biased to blue tone. The color offset can be measured by the color temperature ($K_i$) of different channels,
%
\begin{equation}
K_{i} = \frac{\bar{C}_{\text{R}}+\bar{C}_{\text{G}_\text{r}}+\bar{C}_{\text{G}_\text{b}}+\bar{C}_{\text{B}}}{4\bar{C}_{i}},
\label{corlor_temperature}
\end{equation}
where $\bar{C_{i}}, i\in \left \{\text{R}, \text{G}_{\text{r}},\text{G}_\text{b}, \text{B} \right \} $ represents the average value of channel $i$. In order to obtain the correct color temperature for each channel, we utilize \cite{brooks2018unprocessing} to convert the original RGB frame into raw frame and obtain its original color temperature $\hat{K_{i}}$. Therefore, we can obtain the correction coefficients ($\alpha_{i}$) of the blue and green channels based on the red channel by solving the following equations:
\begin{equation}
\frac{\bar{C}_{\text{R}}+\alpha_{\text{G}_\text{r}}\bar{C}_{\text{G}_\text{r}}+\alpha_{\text{G}_\text{b}}\bar{C}_{\text{G}_\text{b}}+\alpha_{\text{B}}\bar{C}_{\text{B}}}{4\bar{C}_{i}} = \hat{K_{i}}, i\in \left \{ \text{G}_{\text{r}},\text{G}_{\text{b}},\text{B} \right \}.
\label{corlor_temperature}
\end{equation}
The corrected noisy frame can be obtained by
${\hat{\mathbf{I}}^\text{n}_i} = \alpha_i \mathbf{I}^\text{n}_i$, $i\in \left \{ \text{G}_{\text{r}},\text{G}_{\text{b}},\text{B} \right \}$. The corresponding clean frame is also processed with the same $\alpha_i$. Note that, our color calibration coefficients can be treated as linear digital gains. After color correction, the noise distributions in our dataset are close to real normal noisy videos.

In summary, after intensity correction, spatial alignment, and color correction, we obtain the paired noisy-clean frames for supervised learning. For brevity, we still utilize $\mathbf{I}^\text{n}$ to denote the noisy frames in the following.


\begin{figure*}
    \centering
    \includegraphics[width=0.8\linewidth]{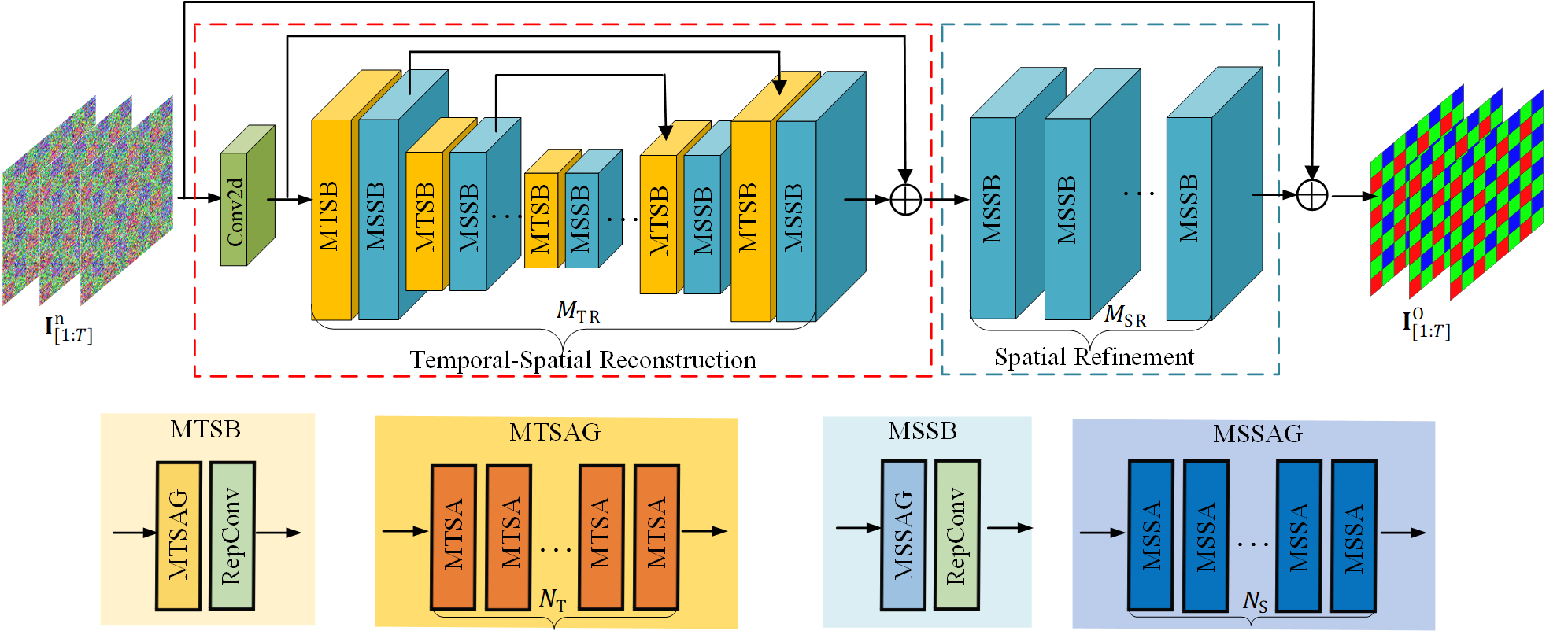}
    \caption{The framework of proposed RViDeformer, which is constructed by temporal-spatial reconstruction (containing $M_{\text{TR}}$ blocks) and spatial refinement (containing $M_{\text{SR}}$ blocks) modules.}
    \label{fig:framework}
\end{figure*}

\section{The Proposed RViDeformer}

In this section, we first introduce the network structure of our proposed raw video denoising transformer (RViDeformer), and then present its key components, \textit{i.e.}, the multi-branch spatial self-attention block (MSSB) and temporal self-attention block (MTSB).

\subsection{Network Structure Overview}
Given a set of consecutive Bayer raw noisy frames $\mathbf{I}^\text{n}_{[1:T]}$, we aim to recover the raw clean frames $\mathbf{I}^{\text{GT}}_{[1:T]}$ through our network RViDeformer, as illustrated in Fig. \ref{fig:framework}. The noisy sequence $\mathbf{I}^\text{n}_{[1:T]}$ is first packed into four channels, and goes through the MTSB and MSSB block alternatively to exploit temporal-spatial correlations. After temporal-spatial reconstruction, the features are fed into the spatial refinement module to generate the denoised raw frames $\mathbf{I}^\text{O}_{[1:T]}$, which are then transformed to sRGB domain via a pre-trained ISP module. There are $M_{\text{TR}}$ blocks in the temporal-spatial reconstruction module and $M_{\text{SR}}$ blocks in the spatial refinement module. The crucial components of our network are the MSSB and MTSB blocks, which are described in detail in the following.

\subsection{Multi-Branch Spatial Self-Attention (MSSA)}
The recent transformer-based video denoising methods \cite{song2022tempformer,liang2024vrt,liang2022recurrent} are all based on Swin Transformer \cite{liu2021swin,liang2021swinir}, which divides images with non-overlapping windows and only apply self-attention inside the window. Although Swin Transformer tries to exchange the information between windows through window shift operation, but the exchanged areas are limited and the long-range information can not be utilized. To solve this problem, we propose Global-Window and Neighbour-Window Self-Attention to enable exploring correlation from global and the neighbor areas. Since the downsampling operation can reduce noise but preserve the low frequency information, we further propose Low-Resolution Self-Attention to make each window not only learn from itself but also from its downsampled version. Afterward, we combine the multiple self-attention branches via feature fusion.

We would like to point out that multi-branch structure has been applied to build transformer in classification \cite{chen2021crossvit,chen2022mobile} and image super-resolution \cite{zhang2022efficient}. Among them, \cite{chen2021crossvit} utilizes dual branch to combine image tokens of different sizes, \cite{chen2022mobile} designs two parallel branch to combine MobileNet and transformer, and \cite{zhang2022efficient} constructs a multi-branch structure where different branches utilize different window size for self-attention. In contrast, we utilize the same window size for the Queries but utilize different window sizes for Keys and Values. In this way, we can utilize correlated information from different receptive fields for video denoising.


\subsubsection{Shift-Window Self-Attention (SWSA)}
The SWSA is the same as that defined in SwinIR \cite{liang2021swinir}. Given a noisy frame feature $\mathbf{F}\in \mathbb{R}^{H\times W\times C}$, we split it into $\lfloor \frac{H}{h} \rfloor \times \lfloor \frac{W}{w} \rfloor$ windows, where the window size is $h\times w$. For the $i$-th window $\mathbf{F}_i\in \mathbb{R}^{N\times C}$ (where $N=hw$), we project it into query $\mathbf{Q}_i$,  key $\mathbf{K}_i$, and value $\mathbf{V}_i$ by linear projection,
\begin{equation}
\mathbf{Q}_i=\mathbf{F}_i \mathbf{P}^\text{Q},\mathbf{\quad K}_i=\mathbf{F}_i \mathbf{P}^\text{K},\mathbf{\quad V}_i=\mathbf{F}_i \mathbf{P}^\text{V},
\label{Eq:D}
\end{equation}
where $\mathbf{P}^\text{Q}, \mathbf{P}^\text{K}, \mathbf{P}^\text{V}\in \mathbb{R}^{C\times D}$ are projection matrices and $D$ is the channel number of projected features. We use $\mathbf{Q}_i$ to query $\mathbf{K}_i$ in order to generate the attention map $\mathbf{A}_i=\text{SoftMax}(\mathbf{Q}_i (\mathbf{K}_i)^\mathsf{T}/\sqrt{D})\in \mathbb{R}^{N\times N}$, and $\mathbf{A}_i$ is used for weighted sum of $\mathbf{V}_i$, namely $\text{SWSA}(\mathbf{Q}_i,\mathbf{K}_i,\mathbf{V}_i) = \mathbf{A}_i\mathbf{V}_i$. The $\text{SoftMax}$ denotes the row softmax operation. In this way, we generate the enhanced feature $\mathbf{F}^{\text{o}}_i\in \mathbb{R}^{N\times D}$, whose noise is reduced through the weighted average of similar features inside the window itself.

\subsubsection{Global-Window Self-Attention (GWSA)}
In SWSA, we utilize the same window to generate the query, key, and value of the transformer. However, this approach limits the calculation of correlations to only occur within the window. For GWSA, we propose to down sample the whole feature map $\mathbf{F}\in \mathbb{R}^{H\times W\times C}$ to the window size to construct a global window $\mathbf{F}^\text{g}\in \mathbb{R}^{N\times C}$. For the $i$-th window, the queries are obtained by linear projection of $\mathbf{F}_i$ (as defined in SWSA), while the keys and values are obtained by linear projection of $\mathbf{F}^\text{g}$. Namely
\begin{equation}
\mathbf{Q}^\text{g}_i=\mathbf{F}_i \mathbf{P}^\text{Q}_\text{g},\mathbf{\quad K}^\text{g}=\mathbf{F}^\text{g} \mathbf{P}^\text{K}_\text{g},\mathbf{\quad V}^\text{g}=\mathbf{F}^\text{g} \mathbf{P}^\text{V}_\text{g},
\end{equation}
where $\mathbf{P}^\text{Q}_\text{g}, \mathbf{P}^\text{K}_\text{g}, \mathbf{P}^\text{V}_\text{g}\in \mathbb{R}^{C\times D^\text{g}}$ are projection matrices and $D^\text{g}$ is the channel number of projected features. Afterwards, we generate the attention map  $\mathbf{A}^\text{g}_i=\text{SoftMax}(\mathbf{Q}^\text{g}_i (\mathbf{K}^\text{g})^\mathsf{T}/\sqrt{D^\text{g}})\in \mathbb{R}^{N\times N}$ to fuse the values $\mathbf{V}^\text{g}$, resulting in $\mathbf{F}^{\text{og}}_i\in \mathbb{R}^{N\times D^\text{g}}$. In this way, the feature of each local window is predicted by the fusion of global downsampled features.

\begin{figure}
    \centering
    \includegraphics[width=1.0\linewidth]{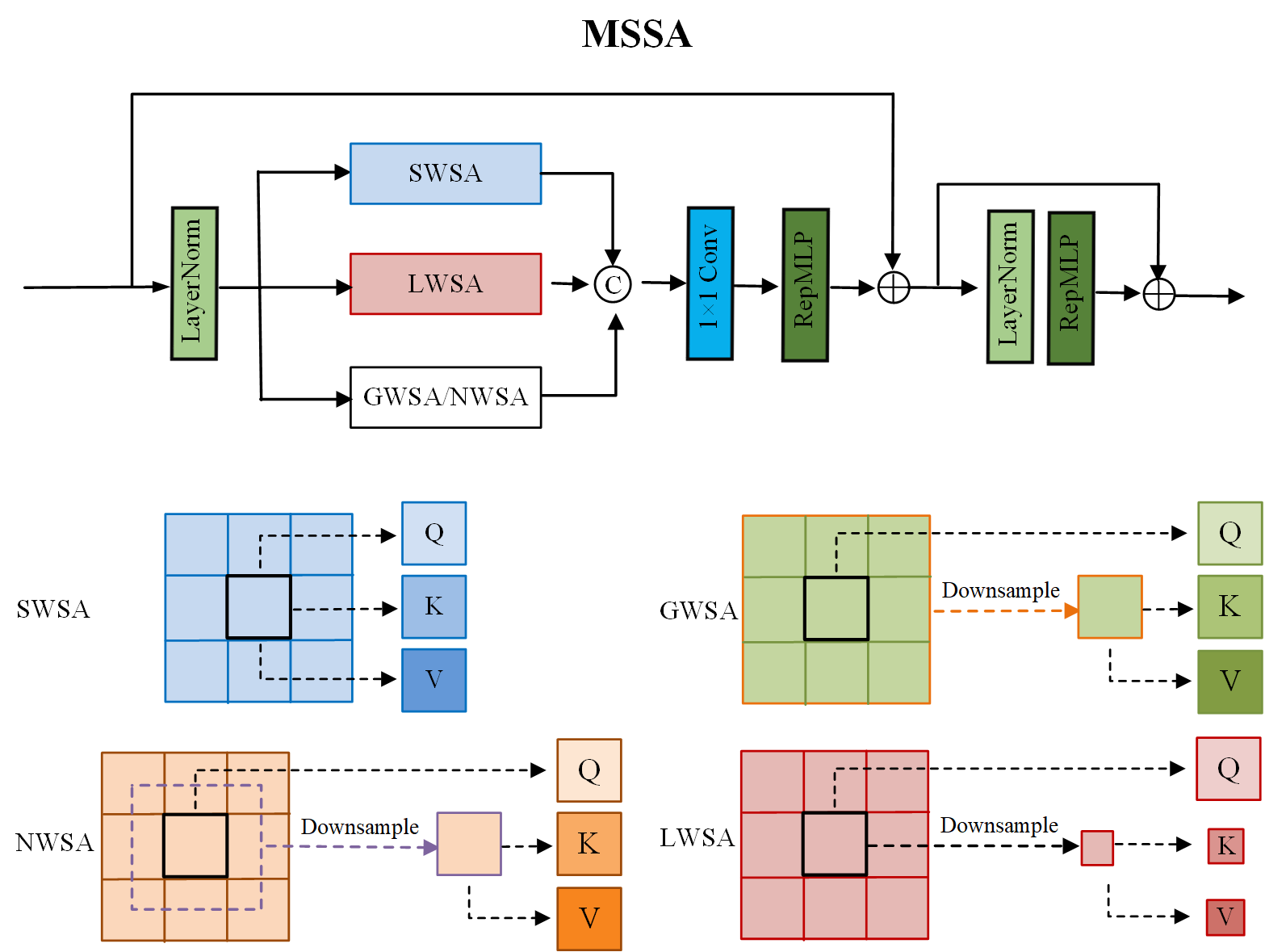}
    \caption{The framework of proposed MSSA.}
    \label{fig:mssa}
\end{figure}

\subsubsection{Neighbour-Window Self-Attention (NWSA)}
When the window number $\lfloor \frac{H}{h} \rfloor \times \lfloor \frac{W}{w} \rfloor$ is large, the global window $\mathbf{F}^\text{g}$ cannot represent the detailed features. Therefore, we further propose Neighbour-Window Self-Attention (NWSA), which utilizes the information from a large neighbour area rather than the global feature map. As shown in Fig. \ref{fig:mssa}, for the $i$-th window $\mathbf{F}_i$, we downsample its neighbour area to make the large neighbor have the same size as $\mathbf{F}_i$, generating a neighbour-window $\mathbf{F}^\text{n}_i\in \mathbb{R}^{h\times w\times C}$, reshaped as $\mathbf{F}^\text{n}_i\in \mathbb{R}^{N\times C}$. Then we compute the query $\mathbf{Q}^\text{n}_i$, key $\mathbf{K}^\text{n}_i$ and value $\mathbf{V}^\text{n}_i$ from $\mathbf{F}_i$ and $\mathbf{F}^\text{n}_i$ by linear projections as
\begin{equation}
\mathbf{Q}^\text{n}_i=\mathbf{F}_i \mathbf{P}^\text{Q}_\text{n},\mathbf{\quad K}^\text{n}_i=\mathbf{F}^\text{n}_i \mathbf{P}^\text{K}_\text{n},\mathbf{\quad V}^\text{n}_i=\mathbf{F}^\text{n}_i \mathbf{P}^\text{V}_\text{n},
\end{equation}
where $\mathbf{P}^\text{Q}_\text{n}, \mathbf{P}^\text{K}_\text{n}, \mathbf{P}^\text{V}_\text{n}\in \mathbb{R}^{C\times D^\text{n}}$ are projection matrices and $D^\text{n}$ is the channel number of projected features. We use $\mathbf{Q}^\text{n}_i$ to query $\mathbf{K}^\text{n}_i$ to generate the attention map $\mathbf{A}^\text{n}_i=\text{SoftMax}(\mathbf{Q}^\text{n}_i (\mathbf{K}^\text{n}_i)^\mathsf{T}/\sqrt{D^\text{n}})\in \mathbb{R}^{N\times N}$, and $\mathbf{A}^\text{n}_i$ is used for weighted sum of $\mathbf{V}^\text{n}_i$, generating the enhanced feature $\mathbf{F}^{\text{on}}_i\in \mathbb{R}^{N\times D^\text{n}}$. In other words, the feature of each local window is predicted by the feature in down-sampled neighbor window. \comment{Note that, when the token is located at the corner or edge of the whole patch, part of the neighbor window will be outside the patch. In this case, we pad these areas with zeros.}

\subsubsection{Low-Resolution-Window Self-Attention (LWSA)}
Multi-scale architecture is beneficial for denoising. In RViDeformer, we not only utilize it in constructing the UNet-like backbone, but also apply it in the transformer structure, namely the values of the transformer are from windows with different receptive fields. For GWSA and NWSA, we are actually utilizing the information from large areas. Therefore, we further propose to utilize the information from the window itself but in a low-resolution version. For the $i$-th window $\mathbf{F}_i$, we downsample it with scale $2$ to construct a lower resolution window $\mathbf{F}^\text{l}_i\in \mathbb{R}^{\frac{N}{4}\times C}$. $\mathbf{F}^\text{l}_i$ reduces the noise in $\mathbf{F}_i$ but still preserves the structure of input image. Then, the query $\mathbf{Q}^\text{l}_i$, key $\mathbf{K}^\text{l}_i$ and value $\mathbf{V}^\text{l}_i$ are derived by
\begin{equation}
\mathbf{Q}^\text{l}_i=\mathbf{F}_i \mathbf{P}^\text{Q}_\text{l},\mathbf{\quad K}^\text{l}_i=\mathbf{F}^\text{l}_i
\mathbf{P}^\text{K}_\text{l},\mathbf{\quad V}^\text{l}_i=\mathbf{F}^\text{l}_i \mathbf{P}^\text{V}_\text{l},
\end{equation}
where $\mathbf{P}^\text{Q}_\text{l}, \mathbf{P}^\text{K}_\text{l}, \mathbf{P}^\text{V}_\text{l}\in \mathbb{R}^{C\times D^\text{l}}$ are projection matrices and $D^\text{l}$ is the channel number of projected features. Then we calculate the attention weights $\mathbf{A}^\text{l}_i=\text{SoftMax}(\mathbf{Q}^\text{l}_i(\mathbf{K}^\text{l}_i)^\mathsf{T}/\sqrt{D^\text{l}})\in \mathbb{R}^{N\times \frac{N}{4}}$ to fuse $\mathbf{V}^\text{l}_i$, generating $\mathbf{F}^{\text{ol}}_i\in \mathbb{R}^{N\times D^\text{l}}$.

As shown in Fig. \ref{fig:mssa}, we construct three self-attention branches, namely SWSA, LWSA, and GWSA (or NWSA) respectively. Specifically, we apply NWSA for the MSSB and MTSB blocks in the original resolution  and GWSA is applied on the other blocks since applying GWSA on the original resolution will lose much information. Since the proposed GWSA, NWSA, and LWSA are also window based, we utilize the shift window operation on them for better performance. Finally, we fuse the three outputs  $\mathbf{F}^{\text{o}}_i, \mathbf{F}^{\text{ol}}_i, \mathbf{F}^{\text{on}}_i(\mathbf{F}^{\text{og}}_i)$ via a 1$\times$1 convolution layer and adjust the contributions of each branch by the parameters $D$, $D^\text{l}$ and $D^\text{n}$ ($D^\text{g}$). We utilize $N_\text{S}$ MSSA to construct one MSSA Group (MSSAG).

Through our proposed GWSA and NWSA, RViDeformer can utilize the information within large windows. In addition, the multi-scale architecture in the temporal-spatial reconstruction implicitly enlarges the window size and helps the information flow from a long distance.

\subsection{Multi-Branch Temporal Mutual Self-Attention (MTSA)}

\begin{figure}
    \centering
    \includegraphics[width=1.0\linewidth]{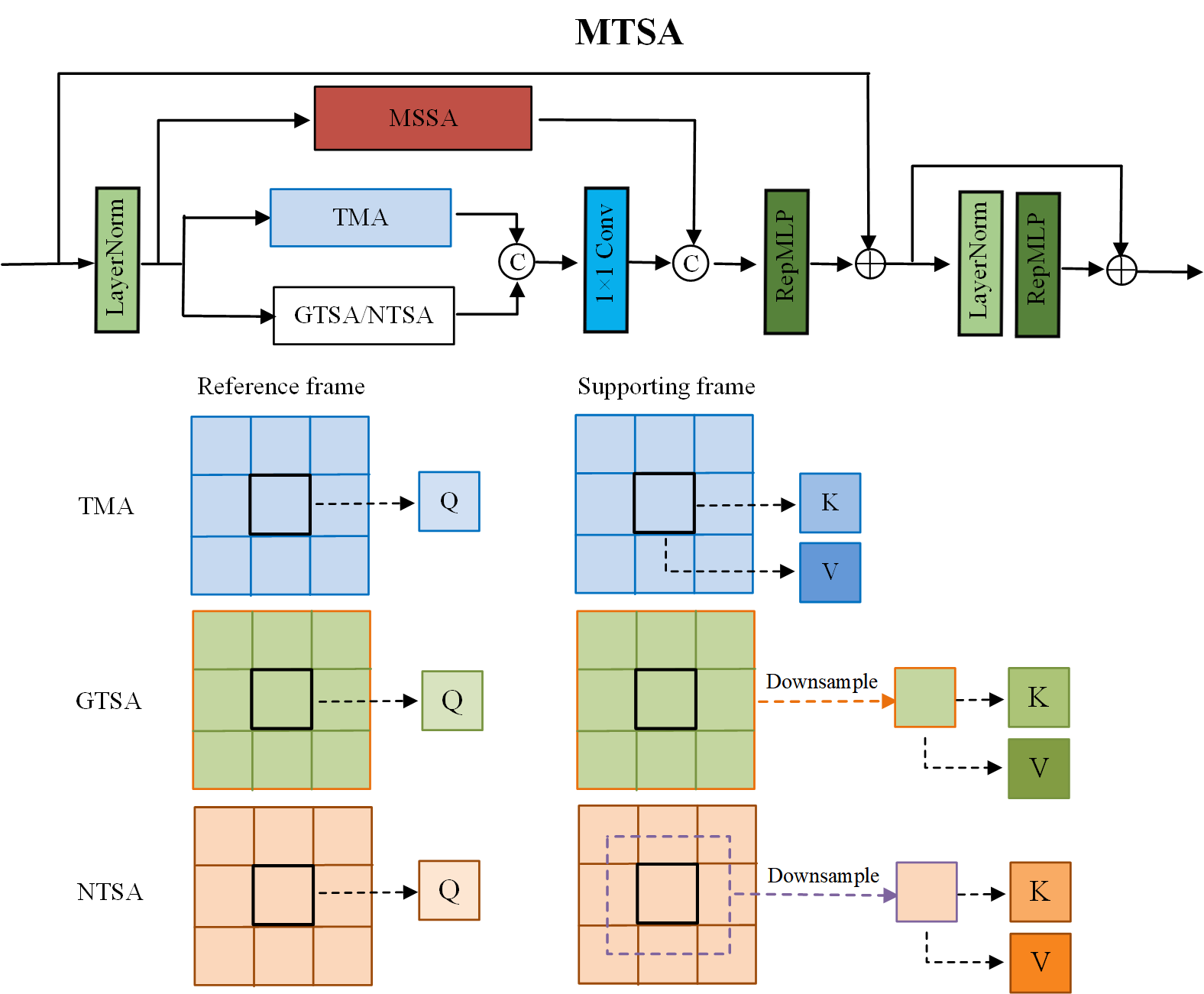}
    \caption{The framework of proposed MTSA.}
    \label{fig:mtsa}
\end{figure}

VRT \cite{liang2024vrt} proposes Temporal Mutual Self-Attention to exploit temporal information from neighbour frames, but the self-attention operation is limited inside the window, which can not process large movements. Therefore, they further introduce warping module to align the neighbor frames (i.e., supporting frames) with the reference frame. Different from it, we propose to utilize transformer to perform implicit alignment by adjusting the window region of the supporting frames. Similar to our proposed MSSA block, we further propose Multi-branch Temporal Mutual Self-Attention (MTSA), which utilizes the Global-Window and Neighbour-Window Self-Attention to process the movements between the reference and supporting frames, as shown in Fig. \ref{fig:mtsa}.

Given a reference frame feature $\mathbf{F}^\text{R}\in \mathbb{R}^{H\times W\times C}$ and a supporting frame feature $\mathbf{F}^\text{S}\in  \mathbb{R}^{H\times W\times C}$, we first split them into windows. For the reference frame, we split it into $\lfloor \frac{H}{h} \rfloor \times \lfloor \frac{W}{w} \rfloor$ windows, where the window size is $h\times w$, and the $i$-th window can be denoted as $\mathbf{F}^\text{R}_i\in \mathbb{R}^{h\times w\times C}$. For the supporting frame, we split it into windows in three ways. The first is the same as that in reference frame, namely the $i$-th window is $\mathbf{F}^\text{S}_i\in \mathbb{R}^{h\times w\times C}$. The second is global-window, namely we directly down-sample the whole feature map to the window size, constructing the global window $\mathbf{F}^{\text{Sg}}\in \mathbb{R}^{h\times w\times C}$. The third is neighbor window, namely we downsample a large neighbor region centered at the $i$-th window into $\mathbf{F}^{\text{Sn}}_i\in  \mathbb{R}^{h\times w\times C}$. According to different window settings in the supporting frame, we construct three different temporal self-attention mechanisms, i.e., the plain Temporal Mutual Attention (TMA), Global-Window Temporal Mutual Attention (GTMA), and Neighbour-Window Temporal Mutual Attention (NTMA). Specifically, the query, key, and value for the three attentions are denoted as
\begin{equation}
\begin{split}
& \text{TMA:} \; \mathbf{Q}^{\text{R}}_i=\mathbf{F}^\text{R}_i
\mathbf{P}^\text{Q}_{\text{R}},\; \mathbf{K}^{\text{S}}_i=\mathbf{F}^{\text{S}}_i \mathbf{P}^\text{K}_{\text{S}},\; \mathbf{V}^{\text{S}}_i=\mathbf{F}^{\text{S}}_i \mathbf{P}^\text{V}_{\text{S}} \\
&\text{GTSA:} \; \mathbf{Q}^{\text{Rg}}_i=\mathbf{F}^{\text{R}}_i \mathbf{P}^\text{Q}_{\text{Rg}},\; \mathbf{K}^{\text{Sg}}=\mathbf{F}^{\text{Sg}}
\mathbf{P}^\text{K}_{\text{Sg}},\; \mathbf{V}^{\text{Sg}}=\mathbf{F}^{\text{Sg}} \mathbf{P}^\text{V}_{\text{Sg}} \\
& \text{NTSA:} \; \mathbf{Q}^{\text{Rn}}_i=\mathbf{F}^{\text{R}}_i \mathbf{P}^\text{Q}_{\text{Rn}},\; \mathbf{K}^{\text{Sn}}_i=\mathbf{F}^{\text{Sn}}_i \mathbf{P}^\text{K}_{\text{Sn}},\; \mathbf{V}^{\text{Sn}}_i=\mathbf{F}^{\text{Sn}}_i \mathbf{P}^\text{V}_{\text{Sn}},\\
\end{split}
\label{eq:qkv}
\end{equation}
where $\mathbf{P}^\text{Q}_{\text{R}}, \mathbf{P}^\text{K}_{\text{S}}, \mathbf{P}^\text{V}_{\text{S}}\in \mathbb{R}^{C\times D_\text{T}}$ , $\mathbf{P}^\text{Q}_{\text{Rg}}, \mathbf{P}^\text{K}_{\text{Sg}}, \mathbf{P}^\text{V}_{\text{Sg}}\in \mathbb{R}^{C\times D^\text{g}_\text{T}}$ and $\mathbf{P}^\text{Q}_{\text{Rn}}, \mathbf{P}^\text{K}_{\text{Sn}}, \mathbf{P}^\text{V}_{\text{Sn}}\in \mathbb{R}^{C\times D^\text{n}_\text{T}}$ are projection matrices. Note that all the features are reshaped into $N\times C$ before the projection operation. Then, we calculate the attention coefficients between the query and corresponding key, and then generate  the fusion result by weighted average of the corresponding values based on the attention coefficients, similar to that in MSSA.

As shown in Fig. \ref{fig:mtsa}, we construct the first branch of MTSA with TMA, and construct the second branch with GTMA (NTMA). Similar to MSSA, NTMA is applied on the blocks with original resolution and GTMA is applied on the other blocks with low resolution. Through our proposed GTMA and NTMA, RViDeformer can utilize the information with large movements from the supporting frames. Note that, we also introduce the MSSA module during temporal reconstruction to further enhance the spatial correlations. We utilize $N_\text{T}$ MTSA to construct one MTSA Group (MTSAG).

Through our proposed GTMA and NTMA, RViDeformer can utilize the information with large movements from the supporting frame. In addition, the multi-scale architecture in the temporal-spatial reconstruction implicitly enlarges the window size and helps the information flow from a long distance. These strategies help our model to deal with videos with large motions.

\comment{The spatial self-attention and temporal mutual self-attention modules follow the window shift operation of SWSA. After the shift operation, original window may be composed of several sub-windows that are not adjacent. In this case, we utilize the original mask mechanism in SWSA to limit self-attention computation to within each sub-window or between the sub-window and its corresponding global window, neighboring window, or low-resolution window.}

\subsection{Reparameterization}

\textbf{Reparameterized MLP (RepMLP).}
For transformer, we usually utilize MLP after the self-attention layers. In this work, we propose to utilize two MLPs during training to increase the capability of the network. During test, we utilize reparameterization to fuse the two linear layers into one linear layer. In this way, the inference cost is the same as that for one MLP layer. Specifically, during training, the $C_{\text{in}}$-channel feature map $\mathbf{I}\in\mathbb{R}^{C_{\text{in}}\times H\times W}$ goes through two parallel linear layers with weights $\mathbf{W}_{\text{L1}}, \mathbf{W}_{\text{L2}}\in \mathbb{R}^{C_{\text{out}}\times C_{\text{in}}}$ and bias $\textbf{\textit{b}}_{\text{L1}}, \textbf{\textit{b}}_{\text{L2}} \in\mathbb{R}^{C_{\text{out}}}$, generating the corresponding $C_{\text{out}}$-channel feature maps $\mathbf{O}_{\text{L1}}$ and $\mathbf{O}_{\text{L2}}\in\mathbb{R}^{C_{\text{out}}\times H\times W}$, respectively. $\mathbf{O}_{\text{L1}}$ and $\mathbf{O}_{\text{L2}}$ are then added together and activated by GELU layer, then goes through a dropout and a linear layer to generate the final result. When inference, based on the linearity of the linear layer, two parallel linear layers can be fused into one linear layer with weights $\mathbf{W}_{\text{Lf}}$ and $\textbf{\textit{b}}_{\text{Lf}}$, which can be formulated as
\begin{equation}
\begin{split}
\mathbf{W}_{\text{Lf}} &= \mathbf{W}_{\text{L1}}+\mathbf{W}_{\text{L2}}, \\
\textbf{\textit{b}}_{\text{Lf}} &= \textbf{\textit{b}}_{\text{L1}}+\textbf{\textit{b}}_{\text{L2}}. \\
\end{split}
\label{eq:fuse1}
\end{equation}

\textbf{Reparameterized Convolution (RepConv).}
In the end of each MTSB (MSSB), we further utilize a 3$\times$3 2D convolution to model the local spatial context. Therefore, the last linear layer (which is equal to a 1$\times$1 2D convolution) in the transformer (MTSAG or MSSAG) can be fused with the convolution layer during inference.
The weights of two convolutions before fusion can be denoted as $\mathbf{W}_{\text{C1}}\in \mathbb{R}^{C_{\text{out}}\times C_{\text{in}}\times1\times1}$, $\mathbf{W}_{\text{C2}}\in \mathbb{R}^{C_{\text{out}}\times C_{\text{in}}\times3\times3}$. Similarly, the bias can be denoted as $\textbf{\textit{b}}_{\text{C1}}, \textbf{\textit{b}}_{\text{C2}} \in\mathbb{R}^{C_{\text{out}}}$. The weight and bias of fused convolution can be denoted as $\mathbf{W}_{\text{Cf}}\in \mathbb{R}^{C_{\text{out}}\times C_{\text{in}}\times3\times3}$ and $\textbf{\textit{b}}_{\text{Cf}}\in\mathbb{R}^{C_{\text{out}}}$. According to \cite{ding2021diverse}, the fused convolution parameters are
\begin{equation}
\begin{split}
\mathbf{W}_{\text{Cf}} &= \text{conv}(\mathbf{W}_{\text{C2}}, \mathbf{W}^T_{\text{C1}}), \\
\textbf{\textit{b}}_{\text{Cf}} &= \text{sum}(\mathbf{W}_{\text{C2}} \textbf{\textit{b}}_{\text{C1}})+\textbf{\textit{b}}_{\text{C2}}. \\
\end{split}
\label{eq:fuse1}
\end{equation}

\subsection{ISP}\label{Sec:ISP}
We utilize the ISP module proposed by \cite{yue2020supervised} to transfer raw denoising results $\mathbf{I}^\text{O}_{[1:T]}$ to the sRGB domain $\mathbf{S}^\text{O}_{[1:T]}$. The ISP module has a UNet architecture, which is trained with 230 clean raw and sRGB pairs from SID dataset \cite{chen2019learning}. By changing the training pairs, we can simulate ISPs of different cameras. In addition, ISP module can also be replaced by traditional ISP pipelines \cite{zamir2020cycleisp}.

\subsection{Loss Functions}

In this work, we adopt two kinds of loss functions, which are used for supervised and unsupervised video denoising tasks, respectively.

\bt{Supervised loss}
Our supervised loss function includes raw and sRGB domain reconstruction losses, which can be formulated as
\begin{equation}
\begin{split}
\mathcal{L}_{\text{raw}}= &\|\mathbf{I}^\text{O}_t-\mathbf{I}^{\text{GT}}_t\|_1 \\
\mathcal{L}_{\text{sRGB}}= &\|\mathbf{S}^\text{O}_t-\mathbf{S}^{\text{GT}}_t\|_1 \\
\mathcal{L}_{\text{sup}}= &\mathcal{L}_{\text{raw}}+\beta_1 \mathcal{L}_{\text{sRGB}} \\
\end{split}
\label{eq:suploss}
\end{equation}
where $\mathbf{I}^{\text{O}}_t$ and $\mathbf{S}^{\text{O}}_t$ denote the raw and sRGB output of the network for the $t$-th frame, $\mathbf{I}^{\text{GT}}_t$ and $\mathbf{S}^{\text{GT}}_t$ denote the corresponding ground truths. The parameters of the pretrained ISP are fixed when training the denoising network, which is beneficial for improving the reconstruction quality in the sRGB domain. $\beta_1$ is the hyper-parameter to balance the two losses.

\bt{Unsupervised loss}
For unsupervised video denoising, the key is to build noisy-noisy pairs for Noise2Noise training \cite{lehtinen2018noise2noise} from video data.
F2F \cite{ehret2019model} and MF2F \cite{dewil2021self} utilize neighboring frames after warping for Noise2Noise training, where the warping error negatively impact the performance.
Therefore we construct noisy-noisy pairs only from single frame. In this work, we utilize the NBR2NBR loss  \cite{huang2021neighbor2neighbor}.
%
%
Specifically, for the $t$-th noisy frame $\mathbf{I}^\text{n}_t$, we sub-sample it with a neighbor down-sampler to get sub-frames $\mathbf{I}^{\text{ns1}}_t$ and $\mathbf{I}^{\text{ns2}}_t$.
We feed the network with $\mathbf{I}^{\text{ns1}}_t$ to generate the denoising result $\mathbf{I}^{\text{sO1}}_t$. Then, with $\mathbf{I}^\text{n}_t$ as input, we get $\mathbf{I}^{\text{O}}_t$ and downsample $\mathbf{I}^{\text{O}}_t$ with the same neighbor down-sampler to get sub-frames $\mathbf{I}^{\text{Os1}}_t$ and $\mathbf{I}^{\text{Os2}}_t$. The spatial NBR2NBR loss can be formulated as
\begin{equation}
\begin{split}
\mathcal{L}_{\text{rec}}= &\|\mathbf{I}^{\text{sO1}}_t-\mathbf{I}^{\text{ns2}}_t\|^2_2 \\
\mathcal{L}_{\text{reg}}= &\|\mathbf{I}^{\text{sO1}}_t-\mathbf{I}^{\text{ns2}}_t-(\mathbf{I}^{\text{Os1}}_t-\mathbf{I}^{\text{Os2}}_t)\|^2_2 \\
\mathcal{L}_{\text{unsup}}= &\mathcal{L}_{\text{rec}}+\beta_2 \mathcal{L}_{\text{reg}}, \\
\end{split}
\label{eq:unsuploss}
\end{equation}
where $\beta_2$ is the hyper-parameter controlling the strength of the $\mathcal{L}_{\text{reg}}$.

\section{Experiments}

\subsection{Implementation Details}
For a fair comparison with other video denoising methods under similar computation cost, we build four different versions of RViDeformer as shown in Table \ref{tab:modelconfig}: RViDeformer-T (Tiny), RViDeformer-S (Small), RViDeformer-M (Medium), and RViDeformer-L (Large). They are designed by changing the base channel number of projected features in MTSA and MSSA (i.e. $D_\text{T}$ in Eq. \ref{eq:qkv} and $D$ in Eq. \ref{Eq:D}), the block number in temporal-spatial reconstruction and spatial refinement modules (i.e., $M_{\text{TR}}$ and $M_{\text{SR}}$), and the number of self-attention block in MTSAG and MSSAG (i.e., $N_\text{T}$ and $N_\text{S}$). For all the four versions, the head number in multi-head self-attention is set to 6. The channel numbers of projected features in LWSA ($D^\text{l}$), GWSA ($D^\text{g}$), and NWSA ($D^\text{n}$) branches are half of that ($D$) in SWSA branch of MSSA block, respectively. The channel numbers of projected features in GTMA ($D^\text{g}_\text{T}$) and NTMA ($D^\text{n}_\text{T}$) branches are half of that ($D_\text{T}$) in TMA branch of MTSA block, respectively.


Since LLRVD \cite{fu2022low} has not released their dataset, we compare our method with state-of-the-art raw video denoising methods on our proposed ReCRVD dataset and CRVD dataset \cite{yue2020supervised}. For supervised and unsupervised video denoising on ReCRVD dataset, we train our network with 12000 epochs, and the learning rate starts with 1e-4 and drops to 5e-5 and 2e-5 after 2/6 and 5/6 of total epochs. For supervised and unsupervised video denoising on CRVD dataset, we train our network with 30000 epochs, and the learning rate starts with 1e-4 and drops to 5e-5 and 2e-5 after 4/6 and 5/6 of total epochs. The hyper-parameter $\beta_1$ in supervised loss is set to 0.5, and the hyper-parameter $\beta_2$ in unsupervised loss is set to 2$\times$(epoch/(total epoch)). The batch size and patch size are set to 1 and 64, respectively. The temporal number $T$ is set to 6 during training. The experiments are conducted with an NVIDIA 3090 GPU.

\begin{table}[t]
\vspace{-3mm}
\caption{Detailed configurations of different versions of our RViDeformer. The MACs are calculated based on the Bayer raw input with a resolution of 1920$\times$1080.}
\vspace{-5mm}
\begin{center}
\small
\resizebox{0.9\linewidth}{!}{
\setlength{\tabcolsep}{0.3mm}{
\vspace{-0.5mm}
\begin{tabular}{l@{\hspace{6pt}}|c@{\hspace{3pt}}|c|c|c|c|c|c|c}

\toprule
Models & $D_{\text{T}}$ & $D$ & $M_{\text{TR}}$ & $M_{\text{SR}}$ & $N_\text{T}$ & $N_\text{S}$ & heads & GMACs  \\
\midrule
		RViDeformer-T & 24 & 24  & 14 & 2  & 1 & 1  & 6  & 38.81\\
		RViDeformer-S & 24 & 30  & 14 & 3  & 2 & 1  & 6  & 71.30 \\
		RViDeformer-M & 24 & 30  & 14 & 4  & 4 & 2  & 6  & 143.67 \\
		RViDeformer-L & 84 & 108 & 14 & 4  & 4 & 2  & 6  & 1790.86 \\
\bottomrule
\end{tabular}}}
\label{tab:modelconfig}
\end{center}
\end{table}

\subsection{Comparison with State-of-the-art Methods}

\begin{table}[t]
\centering
\caption{Quantitative comparisons of PSNR and SSIM for supervised raw video denoising on the ReCRVD test set. The
GMACs for one frame are calculated based on the Bayer raw input with a resolution of
1920$\times$1080.}
\begin{tabular}{m{2.2cm}m{1.0cm}<{\centering}m{0.9cm}<{\centering}m{1.3cm}<{\centering}m{1.3cm}<{\centering}}
\toprule
Methods                                 &Params(M)     & GMACs & raw  & sRGB \\
\hline
VBM4D \cite{matteo2011video}            &-             & -        & 39.68/0.9475 &  33.65/0.8962 \\
FastDVDnet \cite{tassano2020fastdvdnet} &2.48          & 332.23   & 43.49/0.9806 &  38.87/0.9615 \\
EMVD$^*$ \cite{maggioni2021efficient}   &2.62          & 177.82   & 43.32/0.9794 &  38.60/0.9587 \\
BSVD-32 \cite{qi2022real}               &2.45          & 153.65   & 43.56/0.9807 &  39.06/0.9619 \\
\comment{BRVE$^*$ \cite{zhang2024binarized}}      &\comment{35.89}         & \comment{157.13}   & \comment{43.86/0.9822} &	\comment{39.27/0.9651} \\
RViDeformer-M                           &1.12          & 143.67   & \textbf{43.98/0.9823} &  \textbf{39.57/0.9652} \\
\hline
RViDeNet \cite{yue2020supervised}       &8.57          & 2079.74  & 43.71/0.9811 &  39.09/0.9627 \\
MaskDnGAN \cite{paliwal2021multi}       &2.91          & 3006.49  & 43.98/0.9815 &  39.05/0.9629 \\
FloRNN \cite{li2022unidirectional}      &11.82         & 2316.73  & 44.01/0.9829 &  39.42/0.9672 \\
RVRT$^*$ \cite{liang2022recurrent}      &9.07          & 3152.25  & 43.97/0.9821 &  39.43/0.9648 \\
VRT$^*$ \cite{liang2024vrt}             &7.15          & 3056.48  & 44.07/0.9826 &  39.69/0.9662 \\
\comment{ShiftNet$^*$ \cite{li2023simple}}        &\comment{6.93}          & \comment{2350.27}  & \comment{44.22/0.9833} &	\comment{39.91/0.9682} \\
RViDeformer-L                           &6.77          & 1790.86  & \textbf{44.37/0.9837} &  \textbf{40.15/0.9695} \\
\bottomrule
\end{tabular}
\label{CompareReCRVDSup}
\end{table}

\begin{table}[t]
\centering
\caption{Quantitative comparisons of PSNR and SSIM for supervised raw video denoising on the CRVD test set. The
GMACs for one frame are calculated based on the Bayer raw input with a resolution of 1920$\times$1080.  The results of VBM4D are quoted from \cite{yue2020supervised}. The results of EDVR and FastDVDnet are quoted from \cite{maggioni2021efficient}. The results of FastDVDnet-S are quoted from \cite{ostrowski2022bp}.}
\begin{tabular}{m{2.2cm}m{1.0cm}<{\centering}m{0.9cm}<{\centering}m{1.3cm}<{\centering}m{1.3cm}<{\centering}}
\toprule
Methods                                        &Params(M)  & GMACs & raw  & sRGB \\
\hline
VBM4D \cite{matteo2011video}                   &-              & -        & -            &  34.16/0.9220 \\
EMVD \cite{maggioni2021efficient}              &-              & 39.76    & 44.05/0.9890 &  39.53/0.9796 \\
BSVD-16 \cite{qi2022real}                      &-              & 39.38    & 44.10/0.9884 &  40.17/0.9804 \\
RViDeformer-T                                  &0.43           & 38.81    & \textbf{44.34/0.9887} &  \textbf{40.60/0.9813} \\
\hline
FastDVDnet-S \cite{tassano2020fastdvdnet}      &-              & 146.99   & 44.25/0.9887 &  -  \\
BSVD-24 \cite{qi2022real}                      &-              & 87.73    & 44.39/0.9894 &  40.48/0.9820 \\
BP-EVD \cite{ostrowski2022bp}                  &2.56           & 72.39    & 44.42/0.9889 &  - \\
RViDeformer-S                                  &0.63           & 71.30    & \textbf{44.65/0.9894} &  \textbf{41.00/0.9828} \\
\hline
LLRVD \cite{fu2022low}                         &6.36           & 1724.52  & 44.18/0.9880 &  - \\
EDVR \cite{wang2019edvr}                       &-              & 1544.49  & 44.71/0.9902 &  40.89/0.9838 \\
EMVD-L \cite{maggioni2021efficient}            &-              & 1272.75  & 44.58/0.9899 &  -            \\
FastDVDnet \cite{tassano2020fastdvdnet}        &2.48           & 332.50   & 44.30/0.9891 &  39.91/0.9812 \\
\comment{BRVE$^*$ \cite{zhang2024binarized}}   &\comment{35.89}         & \comment{157.13} &\comment{44.20/0.9885}	&\comment{40.33/0.9811} \\
RViDeformer-M                                  &1.12           & 143.67   & \textbf{44.89/0.9901} &  \textbf{41.29/0.9840} \\
\hline
MaskDnGAN \cite{paliwal2021multi}              &2.91           & 3006.49  & 43.04/0.9756 &  39.62/0.9640 \\
RViDeNet \cite{yue2020supervised}              &8.57           & 2079.74  & 43.97/0.9874 &  39.95/0.9792 \\
FloRNN \cite{li2022unidirectional}             &11.82          & 2316.73  & 45.16/0.9907 &  41.01/0.9843 \\
\comment{ShiftNet$^*$ \cite{li2023simple}}     &\comment{6.93}          & \comment{2350.27} &\comment{44.82/0.9900}	&\comment{40.91/0.9836} \\
RViDeformer-L                                  &6.77           & 1790.86  & \textbf{45.45/0.9913} &  \textbf{41.86/0.9860} \\
\bottomrule
\end{tabular}
\label{CompareCRVDInSup}
\end{table}

\begin{figure*}
    \centering
    \includegraphics[width=1.0\linewidth]{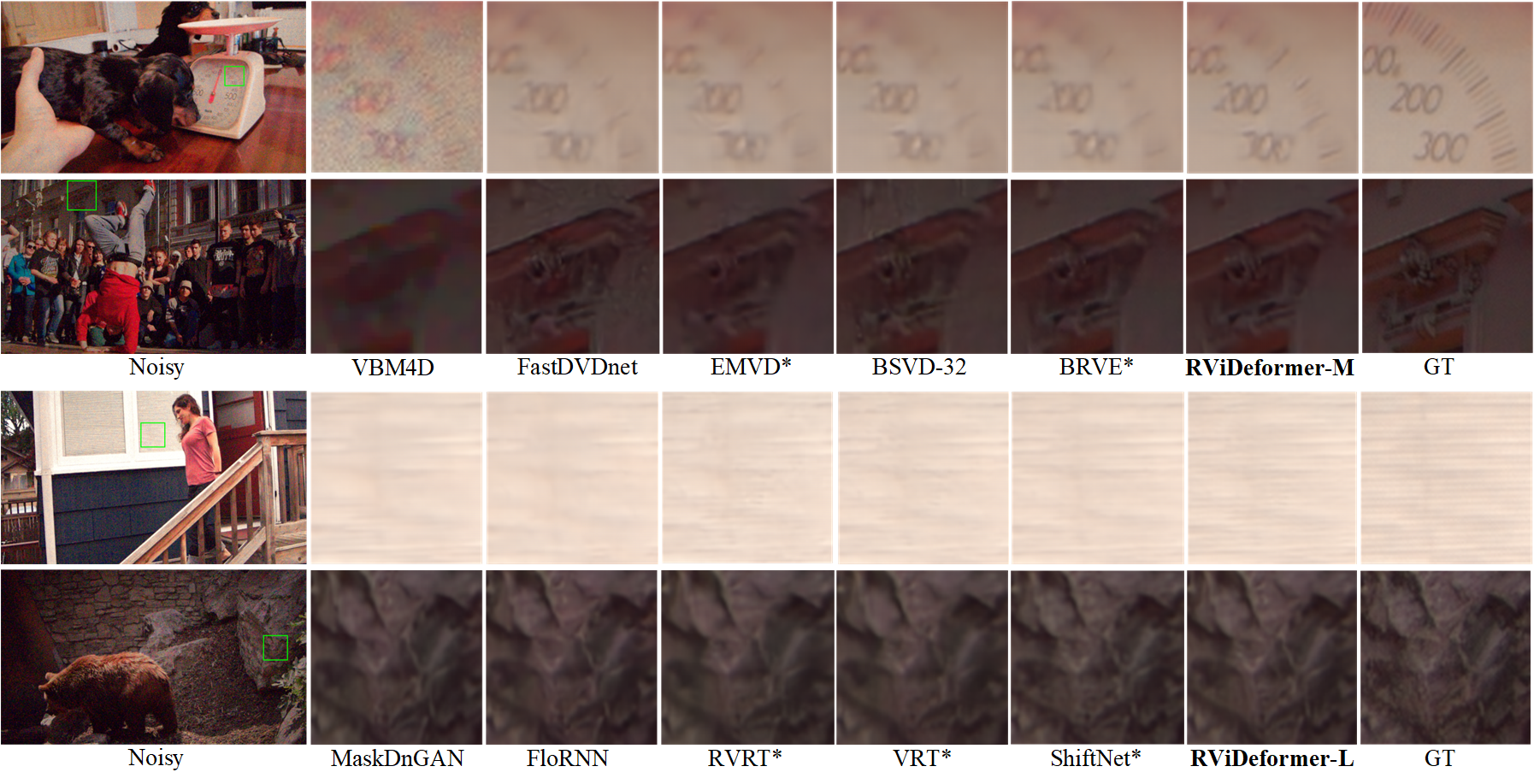}
    \caption{\comment{Visual quality comparison on ReCRVD test set for supervised raw video denoising}. Zoom in for better observation.}
    \label{fig:CompareReCRVDSup}
\end{figure*}

To demonstrate the effectiveness of the proposed raw video denoising methods, we compare with state-of-the-art video denoising methods for supervised learning and unsupervised learning, respectively.

\bt{Supervised learning on ReCRVD dataset} \comment{For supervised learning, we compare our method with eleven video denoising methods, i.e., VBM4D \cite{matteo2011video}, FastDVDnet \cite{tassano2020fastdvdnet}, RViDeNet \cite{yue2020supervised}, MaskDnGAN \cite{paliwal2021multi}, EMVD \cite{maggioni2021efficient}, BSVD \cite{qi2022real}, FloRNN \cite{li2022unidirectional}, RVRT \cite{liang2022recurrent}, VRT \cite{liang2024vrt}, ShiftNet \cite{li2023simple} and BRVE \cite{zhang2024binarized}}. For a fair comparison, all the methods are retrained on the training set of ReCRVD. Due to the significant variation in computation costs among these video denoising methods, we categorize them into two groups based on their computation costs.
The first group is constructed by lightweight video denoising methods (FastDVDnet, EMVD, BSVD-32, \comment{BRVE}) and the traditional denoising method VBM4D. Among them, BSVD-32 is a lighter version of BSVD. Since the code of EMVD has not been released, we use an unofficial implementation \footnote{https://github.com/Baymax-chen/EMVD}, and we denote it as EMVD$^*$. \comment{To unify the computation cost of the first group to a similar level, we increase the base channel number of BRVE for a fair comparison, which is denoted as BRVE$^*$.} For VBM4D, we observe that processing in raw domain can achieve better results. Therefore, we pack the Bayer raw input into 4 channels and process each channel with VBM4D separately. We compare these light-weight video denoising methods with RViDeformer-M, which is a lighter version of our network. The second group contains large video denoising models (RViDeNet, MaskDnGAN, FloRNN, RVRT, VRT, \comment{ShiftNet}) with more computation costs, and we compare them with our RViDeformer-L. Since the optical flow network in VRT and RVRT can only process 3-channel RGB inputs and can not process raw data, we apply a simple ISP to the raw inputs before feeding them to the optical flow network. To unify the computation cost of the second group to a similar level, we reduce the base channel number and the depths of VRT for a fair comparison, which is denoted as VRT$^*$. \comment{For the same reason, we reduce the base channel number of RVRT and ShiftNet, which are denoted as RVRT$^*$ and ShiftNet$^*$, respectively.} We evaluate these methods on the test set of proposed ReCRVD dataset.

Table \ref{CompareReCRVDSup} lists the average PSNR and SSIM values of raw video denoising results in raw and sRGB domains on ReCRVD test set, which contains 30 videos and each video contains 25 frames.
It can be observed that our model greatly outperforms the state-of-the-art small and large video denoising models. \comment{For lighter video denoising methods, RViDeformer-M outperforms the second best method BRVE$^*$ by 0.12 (0.30) dB in raw (sRGB) domain. For heavier video denoising methods, RViDeformer-L outperforms the second best method ShiftNet$^*$  0.15 (0.24) dB in raw (sRGB) domain.} In addition, our model consumes the lowest MACs  (multiply-add operations). In terms of parameters, RViDeformer-M has fewer parameters than other lighter video denoising methods, even fewer than heavier methods RViDeNet, MaskDnGAN, and RVRT$^*$, but it achieves better PSNR and SSIM. RViDeformer-L also has fewer parameters compared to other heavier video denoising methods except for MaskDnGAN. Fig. \ref{fig:CompareReCRVDSup} presents the visual comparison results on four scenes of ReCRVD test set. It can be observed
that our method can remove the noise clearly and recover the most details. VBM4D either cannot remove the noise (such as the first scene) or generate over-smooth results (such as the second scene).
\comment{The results of light-weight methods, i.e., FastDVDnet, EMVD$^*$, BSVD-32 and BRVE$^*$ are a bit smooth in the first scene, and contain artifacts in the second scene.} Compared with computational heavy methods, our method recovers more details, as shown in the third and fourth scenes of Fig. \ref{fig:CompareReCRVDSup}.

Compared with the transformer-based methods RVRT and VRT, our method can utilize long-range information in the frame and handle large movements between frames due to the proposed global and neighbor window self-attention. Thus our method has better quantitative results and restores more details. Benefiting from the transformer architecture, our method performs better than CNN-based methods FastDVDnet, EMVD, BSVD, RViDeNet, MaskDnGAN, FloRNN and ShiftNet. Among them, the GAN-based method MaskDnGAN restores details that result in good visual quality. However, since the generated details deviate from ground truth, it leads to lower PSNR and SSIM values. Our method can restore details with high fidelity. The employed reparameterization in our method further reduces computation costs, resulting in lower MACs and parameters.

\begin{figure*}
    \centering
    \includegraphics[width=1.0\linewidth]{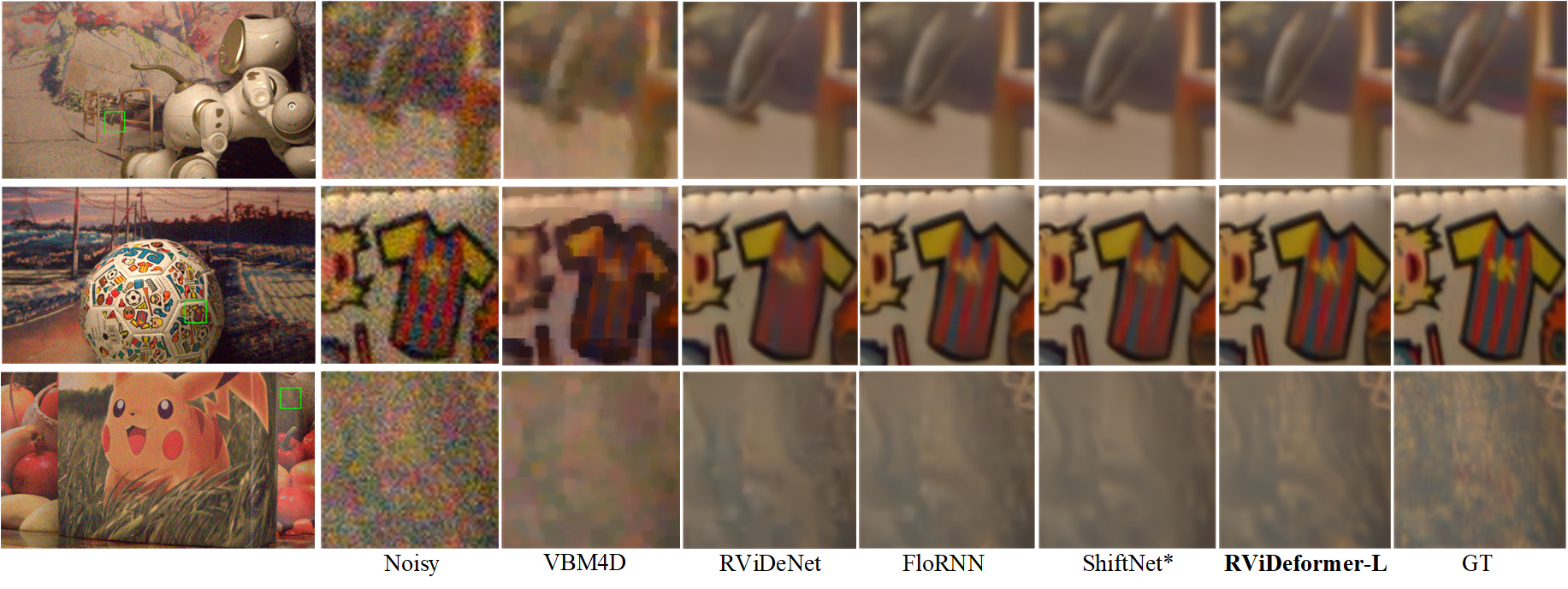}
    \caption{\comment{Visual quality comparison on CRVD  test set for supervised raw video denoising}. Zoom in for better observation.}
    \label{fig:CompareCRVDSup}
\end{figure*}

\begin{figure*}
    \centering
    \includegraphics[width=1.0\linewidth]{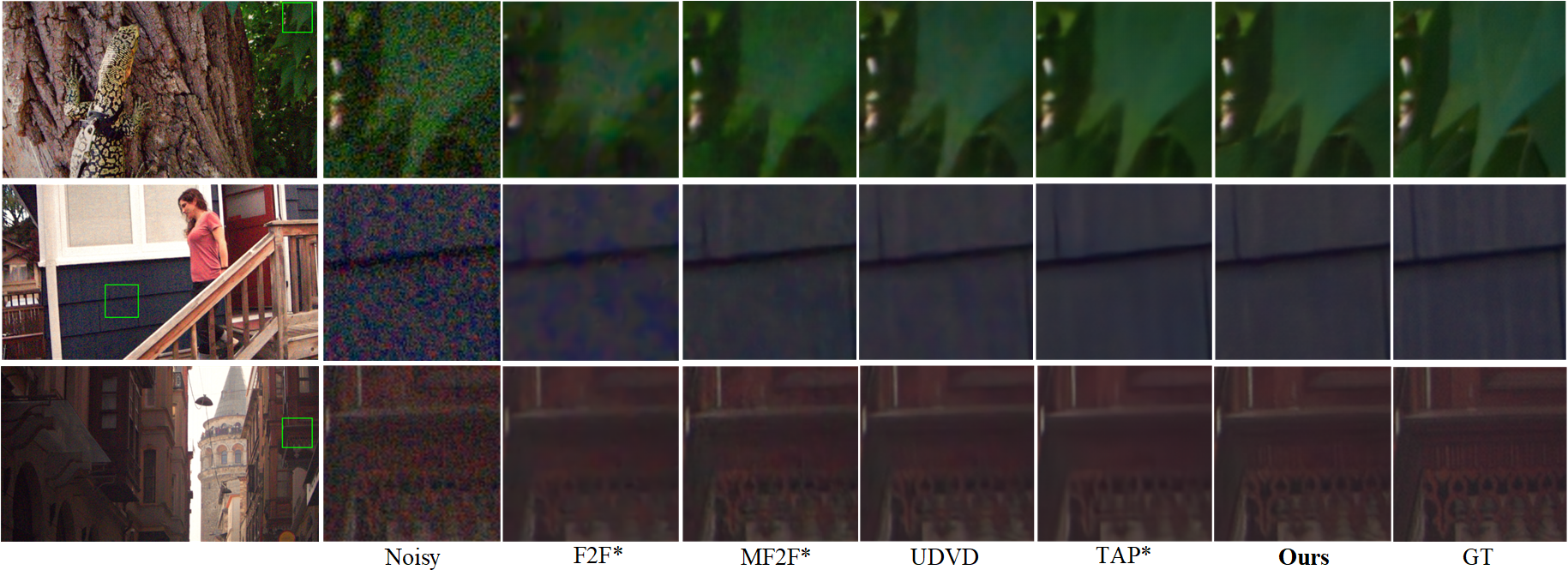}
    \caption{\comment{Visual quality comparison on ReCRVD test set for unsupervised raw video denoising}. Zoom in for better observation.}
    \label{fig:CompareReCRVDUnsup}
\end{figure*}

\bt{Supervised learning on CRVD indoor dataset} We also give the comparison results on CRVD indoor dataset \cite{yue2020supervised}. CRVD indoor dataset is a popular benchmark dataset for raw video denoising, which contains 55 videos among which 6 scenes are used for training and the other 5 scenes are used for testing. Most of the above mentioned methods have been evaluated on this dataset. Therefore, we directly quote their scores from their papers. We further introduce BP-EVD \cite{ostrowski2022bp} and LLRVD \cite{fu2022low} for comparison. Since they are not open-sourced, we did not compare them on ReCRVD dataset. \comment{We also retrain and compare BRVE$^*$ and ShiftNet$^*$ on CRVD dataset, which have been ranked as the second best methods on ReCRVD dataset.} Since we have compared with RVRT and VRT on ReCRVD dataset, we do not retrain and compare them on CRVD dataset. For the methods with light-weight models, we compare these methods with RViDeformer-S or RViDeformer-T. In summary, we categorize these methods into four groups and the methods in each group have similar computation costs.

Table \ref{CompareCRVDInSup} lists the average PSNR and SSIM values of raw video denoising results in raw and sRGB domains on CRVD test set. It can be observed that our models greatly outperform the SOTA video denoising methods in each group. For the first group, RViDeformer-T outperforms the second best method BSVD-16 by 0.24 (0.43) dB in raw (sRGB) domain. For the second group, RViDeformer-S outperforms the second best method BP-EVD by 0.23 dB in raw domain. \comment{For the third group, RViDeformer-M outperforms the second best method BRVE$^*$ by 0.69 (0.96) dB in raw (sRGB) domain.} For the fourth group, RViDeformer-L outperforms the second best method FloRNN by 0.29 (0.85) dB in raw (sRGB) domain. For each group, our method has the lowest MACs. Fig. \ref{fig:CompareCRVDSup} presents the visual comparison results on three scenes of CRVD indoor test set. Since only RViDeNet and FloRNN have released their models, we only compare them and VBM4D on CRVD dataset. It can be observed that our method recovers the most details in all three scenes.

\bt{Unsupervised learning on ReCRVD dataset} For unsupervised learning, we compare our method with state-of-the-art unsupervised video denoising methods F2F \cite{ehret2019model}, MF2F \cite{dewil2021self}, UDVD \cite{sheth2021unsupervised} and TAP \cite{fu2024temporal}. All methods are retrained on the training set of ReCRVD dataset without utilizing the ground truths. F2F and MF2F utilize DnCNN \cite{zhang2017beyond} and FastDVDNet as their denoising network, respectively. We utilize RViDeformer-L as our denoising network. \comment{For a fair comparison, we increase the channel number of DnCNN, FastDVDnet and TAP to unify the computation cost of them to be similar to that of our method, and we denote them as F2F$^*$, MF2F$^*$ and TAP$^*$, respectively. The original TAP utilizes an image denoiser NAFNet with supervised training as backbone. For a fair comparison, we retrained its backbone in an unsupervised manner using the NBR2NBR loss.} We evaluate all methods on the test set of ReCRVD dataset. Table \ref{CompareReCRVDUnsup} lists the average PSNR and SSIM values of unsupervised raw video denoising results on ReCRVD test set in both raw and sRGB domains. It can be observed that our model outperforms the SOTA unsupervised video denoising methods. \comment{Compared with the second and third best methods UDVD and TAP$^*$, our method achieves 0.02 (1.11) and 0.62 (0.86) dB in raw (sRGB) domain, respectively. In addition, our method has much higher SSIM values than UDVD.} The main reason is that UDVD tends to generate smooth results in raw domain (which leads to good PSNR result but worse SSIM value in raw domain) and its sRGB domain results contain many artifacts (which leads to worse PSNR values in sRGB domain). Our method also has the lowest MACs. It proves that our proposed network RViDeformer is also a good architecture for unsupervised video denoising.
Fig. \ref{fig:CompareReCRVDUnsup} presents the visual comparison results on the ReCRVD test set for unsupervised raw video denoising. It can be observed that the results of compared three unsupervised methods contains much chroma noise. The results of F2F$^*$ and MF2F$^*$ contain some artifacts. \comment{Meanwhile, the results of F2F$^*$, UDVD and TAP$^*$ are over-smooth in the third scene.} In contrast, our method removes the noise clearly and recover the most details. 

\textcolor{black}{The main reason for our superior performance is as follows. F2F and MF2F warp the neighboring frames for Noise2Noise training, the warping error limits the performance. UDVD integrates the blind-spot network into the FastDVDnet. Since UDVD applies the same blind spot processing to both the current frame and the neighboring frames, the information of the neighbor frame is lost, which also destroys the utilization of temporal similarity. \comment{TAP generates pseudo labels from its image denoising backbone, which limits the performance.} Different from these methods, we combine NBR2NBR loss with our RViDeformer. The NBR2NBR loss subsamples each frame to construct noisy pairs for Noise2Noise training. The constructed noisy pairs have better quality than the pairs by warping the neighboring frames. Meanwhile, our RViDeformer can still take full advantage of the temporal similarity. Therefore, our method still generates the best results via unsupervised learning.}

\bt{Unsupervised learning on CRVD indoor dataset} We also give the comparison results of unsupervised video denoising methods on CRVD indoor dataset. We retrain the above \comment{four} unsupervised methods on the training set and evaluate them on the test set of CRVD. In UDVD \cite{sheth2021unsupervised}, its results are generated by training and testing on the same scenes of CRVD test set. Since it is inconsistent with the real-world case, we retrain UDVD on the training set and evaluate it with the test set. Table \ref{CompareCRVDInUnsup} lists the average PSNR and SSIM values of unsupervised raw video denoising results on CRVD test set. It can be observed that our model outperforms the second best method TAP* with 0.65 dB gain in sRGB domain. 
Due to page limitations, the visual comparison results are presented in the supplementary material.

To demonstrate the effectiveness of our proposed ReCRVD dataset and raw video denoising method RViDeformer, we use the CRVD outdoor test set \cite{yue2020supervised} for the generalization test. Due to page limitations, the results are presented in the supplementary material.

\begin{table}[t]
\centering
\caption{Quantitative comparisons of PSNR and SSIM for unsupervised raw video denoising on ReCRVD test set. The GMACs for one frame are calculated based on the Bayer raw input with a resolution of 1920$\times$1080.}
\begin{tabular}{m{1.6cm}m{1.1cm}<{\centering}m{0.9cm}<{\centering}m{1.3cm}<{\centering}m{1.3cm}<{\centering}}
\toprule
Methods                              &Params(M)    & GMACs    & raw           & sRGB \\
\hline
F2F$^*$ \cite{ehret2019model}        &4.16       & 2157.57  & 39.00/0.9621  & 32.93/0.9224 \\
MF2F$^*$ \cite{dewil2021self}        &21.9       & 2584.41  & 41.83/0.9726  & 36.15/0.9442 \\
UDVD \cite{sheth2021unsupervised}    &2.98       & 17077.07 & 43.11/0.9785  & 37.87/0.9560 \\
\comment{TAP$^*$ \cite{fu2024temporal}}        &\comment{60.18}      & \comment{2689.45}	& \comment{42.51/0.9780}	& \comment{38.12/0.9586} \\
Ours                                 &6.77       & 1790.86  & \textbf{43.13/0.9810}  & \textbf{38.98/0.9639} \\
\bottomrule
\end{tabular}
\label{CompareReCRVDUnsup}
\end{table}

\begin{table}[t]
\centering
\caption{Quantitative comparisons of PSNR and SSIM for unsupervised raw video denoising on the CRVD test set. The
GMACs for one frame are calculated based on the Bayer raw input with a resolution of 1920$\times$1080. }
\begin{tabular}{m{1.6cm}m{1.1cm}<{\centering}m{0.9cm}<{\centering}m{1.3cm}<{\centering}m{1.3cm}<{\centering}}
\toprule
Methods                                &Params(M)  & GMACs    & raw           & sRGB\\
\hline
F2F$^*$ \cite{ehret2019model}          &4.16     & 2157.57  & 35.31/0.9516 & 27.49/0.9099 \\
MF2F$^*$ \cite{dewil2021self}          &21.9     & 2584.41  & 41.24/0.9774 & 36.95/0.9607 \\
UDVD \cite{sheth2021unsupervised}      &2.98     & 17077.07 & 43.08/0.9851 & 38.71/0.9755 \\
\comment{TAP$^*$ \cite{fu2024temporal}}        &\comment{60.18}      & \comment{2689.45} &\comment{43.11/0.9861} &\comment{39.62/0.9775} \\
Ours                                   &6.77     & 1790.86  & \textbf{43.12/0.9879} & \textbf{40.27/0.9812} \\
\bottomrule
\end{tabular}
\label{CompareCRVDInUnsup}
\end{table}

\subsection{Ablation Study}

\begin{table}[t]
\centering
\caption{Ablation study for the MTSB, MSSB block and the multi-branch self-attention in RViDeformer-M by evaluating on ReCRVD test set.}
\resizebox{0.45\textwidth}{33mm}{
\begin{tabular}{cccccc}
\toprule
\multirow{3}{*}{Block}              & MTSB                           & $\times$      & $\checkmark$  & $\times$     & $\checkmark$   \\
\multirow{3}{*}{}                   & MSSB                           & $\times$      & $\times$      & $\checkmark$  & $\checkmark$   \\
\multirow{3}{*}{}                   & raw PSNR                       & 43.58   & 43.82   & 43.86    & 43.98 \\
\multirow{3}{*}{}                   & raw SSIM                       & 0.9804  & 0.9816  & 0.9616   & 0.9823 \\
\multirow{3}{*}{}                   & sRGB PSNR                      & 39.08   & 39.34   & 39.39    & 39.57 \\
\multirow{3}{*}{}                   & sRGB SSIM                      & 0.9613  & 0.9631  & 0.9636    & 0.9652\\
\midrule
\multirow{3}{*}{MTSB}               & GTMA                           & $\times$      & $\checkmark$  & $\checkmark$ \\
\multirow{3}{*}{}                   & NTMA                           & $\times$      & $\times$      & $\checkmark$ \\
\multirow{3}{*}{}                   & raw PSNR                       & 43.74         & 43.79          & 43.82 \\
\multirow{3}{*}{}                   & raw SSIM                       & 0.9812        & 0.9814         & 0.9816 \\
\multirow{3}{*}{}                   & sRGB PSNR                      & 39.24         & 39.30          & 39.34 \\
\multirow{3}{*}{}                   & sRGB SSIM                      & 0.9627        & 0.9629         & 0.9631 \\
\midrule
\multirow{3}{*}{MSSB}               & LWSA                           & $\times$      & $\checkmark$  & $\checkmark$   & $\checkmark$ \\
\multirow{3}{*}{}                   & GWSA                           & $\times$      & $\times$      & $\checkmark$   & $\checkmark$ \\
\multirow{3}{*}{}                   & NWSA                           & $\times$      & $\times$      & $\times$       & $\checkmark$ \\
\multirow{3}{*}{}                   & raw PSNR                       & 43.71         & 43.76         & 43.82          & 43.86 \\
\multirow{3}{*}{}                   & raw SSIM                       & 0.9812        & 0.9813        & 0.9815         & 0.9816\\
\multirow{3}{*}{}                   & sRGB PSNR                      & 39.19         & 39.26         & 39.35          & 39.39 \\
\multirow{3}{*}{}                   & sRGB SSIM                      & 0.9627        & 0.9630        & 0.9636         & 0.9636 \\
\bottomrule
\end{tabular}
}
\label{tab:Ablation}
\end{table}

In this section, we perform ablation study to demonstrate the effectiveness of the proposed MTSB, MSSB, and multi-branch self-attention. Table \ref{tab:Ablation} lists the quantitative comparison results on ReCRVD test set by removing these modules one by one from RViDeformer-M. In the first row of Table \ref{tab:Ablation}, we replace the proposed MSSB and MTSB by the existed solution to create the baseline model. Specifically, removing MSSB means that we replace the MSSA module in MSSB by a plain SWSA in VRT \cite{liang2024vrt} and the RepConv module in MSSB is replaced by the linear layer in VRT. Removing MTSB means that we replace MTSA with the temporal mutual self-attention (TMSA) in VRT. Therefore, our model without MSSB and MTSB can be regarded as the VRT \cite{liang2024vrt} model without optical flow. As shown in Table \ref{tab:Ablation}, MSSB brings 0.28 (0.31) dB gain in raw (sRGB) domain. MTSB brings 0.24 (0.26) dB gain in raw (sRGB) domain. It demonstrates that the proposed multi-branch self-attention and RepConv are beneficial for video denoising.


The main component for MTSB is the proposed MTSA module, which contains the plain TMA, GTSA, and NTSA module. Therefore, we further ablate the proposed GTSA and NTSA branch. Since we apply NTSA on the features with original resolution and GTSA is applied on low-resolution features. We replace NTMA with GTMA when removing NTMA. It can be observed that PSNR value in the raw (sRGB) domain is decreased by 0.03 (0.04) dB by removing NTMA. If we remove both GTMA and NTMA, namely that there is only TMA and MSSA branches in MTSA, the performance is decreased by 0.08 and 0.1 dB in raw and sRGB domain, respectively.
The key component of MSSB is the proposed MSSA module, which is constructed by three branches. Similar to NTMA, for the version without NWSA, we utilize GWSA on all the blocks. It can be observed that PSNR value in the raw (sRGB) domain is decreased by 0.04 (0.04) dB by removing NWSA. For the version without GWSA, we remove both GWSA and NWSA, and the PSNR value for this version is decreased by 0.1 (0.13) dB in raw (sRGB) domain. For the version without LWSA, GWSA, and NWSA, namely that we only utilize the plain SWSA branch, the PSNR value is decreased by 0.15 (0.2) dB in raw (sRGB) domain. 
For reparameterization, the computation cost is reduced by 3.37 and 13.92 GMACs for the 1920$\times$1080 Bayer input by introducing the proposed RepMLP and RepConv method during inference, respectively.


\subsection{Limitation}

\begin{figure}
    \centering
    \includegraphics[width=0.8\linewidth]{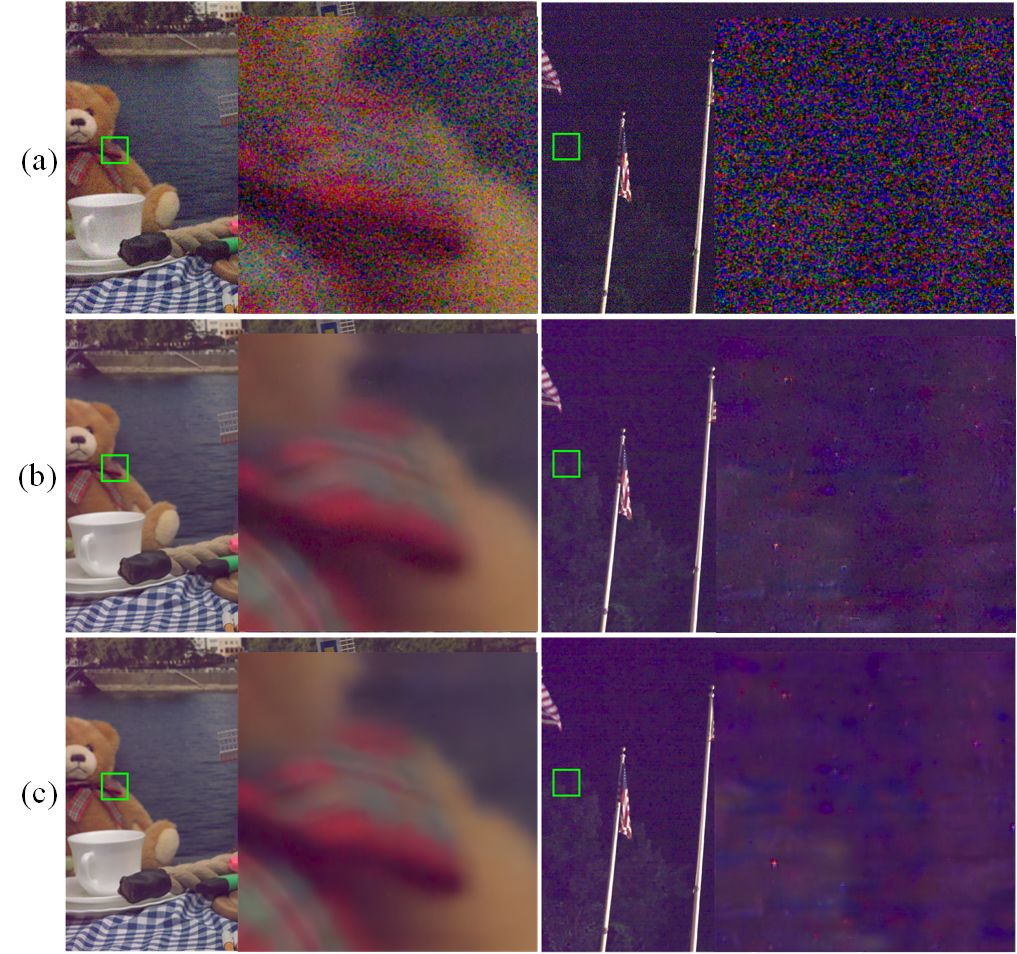}
    \caption{\comment{Failure cases of our method (c) and BRVE* (b). The inputs (a) are from the SMID dataset.}}
    \label{fig:FailureCase}
\end{figure}

\comment{ We would like to point out that our method also has some limitations. Our model is trained with the data captured by sensor IMX385 under five ISO levels. The denoising performance may be degraded when directly applying the model to noisy videos captured by other sensors, or the noisy level is much higher than our ISO settings. Fig. \ref{fig:FailureCase} presents the denoising results of our method (c), namely RViDeformer-M, and BRVE* \cite{zhang2024binarized} (b) on the SMID \cite{chen2019seeing} dataset, which is captured by another sensor. Our model and BRVE* are both trained on ReCRVD dataset. The noise level of the second scene is much higher than that in our dataset. For the first scene our method can remove the noise but the result contains purple-colored artifacts. For the second scene, our method cannot remove the noise clearly and results in severe artifacts. Note that, this is the common limitations of supervised learning, and the second best method BRVE* also generates this kind of artifacts, as shown in Fig. \ref{fig:FailureCase} (b). In the future, we would like to introduce more accurate noise models to enrich our dataset and make it adapt to various sensors.}

\section{Conclusion}

In this work, we construct a benchmark dataset (ReCRVD) for raw video denoising to facilitate futher works on raw video denoising. Compared with CRVD, the motions in ReCRVD are real and the number of scenes (120) is much larger. Correspondingly we propose an efficient raw video denoising transformer network (RViDeformer), which is mainly constructed by multi-branch spatial self-attention and multi-branch temporal self-attention blocks. RViDeformer is evaluated by supervised learning and unsupervised learning, and it achieves the best results on both ReCRVD and CRVD datasets for both the two learning manners. In addition, the models trained on ReCRVD outperforms that trained on CRVD when testing on the third real test set (CRVD outdoor), which demonstrates that the noise model of ReCRVD is consistent with that of real-world noisy videos.

It should be noted that the proposed RViDeformer is not specifically designed for raw data. It can also be applied to sRGB video denoising. \textcolor[rgb]{0.0,0.00,0.00}{In addition, the proposed MSSA and MTSA modules can also be applied to other restoration tasks, such as super-resolution \cite{fang2022cross}  and adverse weather removal \cite{cheng2024progressive,cheng2024continual}, since these tasks can also benefit from a large receptive field. In the future, we would like to extend our work to these restoration tasks.}

%

%

\ifCLASSOPTIONcaptionsoff
  \newpage
\fi



\bibliographystyle{IEEEtran}
\bibliography{egbib}

\begin{thebibliography}{10}
\providecommand{\url}[1]{#1}
\csname url@samestyle\endcsname
\providecommand{\newblock}{\relax}
\providecommand{\bibinfo}[2]{#2}
\providecommand{\BIBentrySTDinterwordspacing}{\spaceskip=0pt\relax}
\providecommand{\BIBentryALTinterwordstretchfactor}{4}
\providecommand{\BIBentryALTinterwordspacing}{\spaceskip=\fontdimen2\font plus
\BIBentryALTinterwordstretchfactor\fontdimen3\font minus
  \fontdimen4\font\relax}
\providecommand{\BIBforeignlanguage}[2]{{%
\expandafter\ifx\csname l@#1\endcsname\relax
\typeout{** WARNING: IEEEtran.bst: No hyphenation pattern has been}%
\typeout{** loaded for the language `#1'. Using the pattern for}%
\typeout{** the default language instead.}%
\else
\language=\csname l@#1\endcsname
\fi
#2}}
\providecommand{\BIBdecl}{\relax}
\BIBdecl

\bibitem{yue2020supervised}
H.~Yue, C.~Cao, L.~Liao, R.~Chu, and J.~Yang, ``Supervised raw video denoising
  with a benchmark dataset on dynamic scenes,'' in \emph{Proceedings of the
  IEEE/CVF Conference on Computer Vision and Pattern Recognition}, 2020, pp.
  2301--2310.

\bibitem{kong2020comprehensive}
Z.~Kong, X.~Yang, and L.~He, ``A comprehensive comparison of multi-dimensional
  image denoising methods,'' \emph{arXiv preprint arXiv:2011.03462}, 2020.

\bibitem{fu2022low}
Y.~Fu, Z.~Wang, T.~Zhang, and J.~Zhang, ``Low-light raw video denoising with a
  high-quality realistic motion dataset,'' \emph{IEEE Transactions on
  Multimedia}, 2022.

\bibitem{song2022tempformer}
M.~Song, Y.~Zhang, and T.~O. Ayd{\i}n, ``Tempformer: Temporally consistent
  transformer for video denoising,'' in \emph{European Conference on Computer
  Vision}.\hskip 1em plus 0.5em minus 0.4em\relax Springer, 2022, pp. 481--496.

\bibitem{liang2024vrt}
J.~Liang, J.~Cao, Y.~Fan, K.~Zhang, R.~Ranjan, Y.~Li, R.~Timofte, and
  L.~Van~Gool, ``Vrt: A video restoration transformer,'' \emph{IEEE
  Transactions on Image Processing}, 2024.

\bibitem{liang2022recurrent}
J.~Liang, Y.~Fan, X.~Xiang, R.~Ranjan, E.~Ilg, S.~Green, J.~Cao, K.~Zhang,
  R.~Timofte, and L.~Van~Gool, ``Recurrent video restoration transformer with
  guided deformable attention,'' \emph{arXiv preprint arXiv:2206.02146}, 2022.

\bibitem{liu2021swin}
Z.~Liu, Y.~Lin, Y.~Cao, H.~Hu, Y.~Wei, Z.~Zhang, S.~Lin, and B.~Guo, ``Swin
  transformer: Hierarchical vision transformer using shifted windows,'' in
  \emph{Proceedings of the IEEE/CVF International Conference on Computer
  Vision}, 2021, pp. 10\,012--10\,022.

\bibitem{liang2021swinir}
J.~Liang, J.~Cao, G.~Sun, K.~Zhang, L.~Van~Gool, and R.~Timofte, ``Swinir:
  Image restoration using swin transformer,'' in \emph{Proceedings of the
  IEEE/CVF international conference on computer vision}, 2021, pp. 1833--1844.

\bibitem{huang2021neighbor2neighbor}
T.~Huang, S.~Li, X.~Jia, H.~Lu, and J.~Liu, ``Neighbor2neighbor:
  Self-supervised denoising from single noisy images,'' in \emph{Proceedings of
  the IEEE/CVF conference on computer vision and pattern recognition}, 2021,
  pp. 14\,781--14\,790.

\bibitem{dong2018denoising}
W.~Dong, P.~Wang, W.~Yin, G.~Shi, F.~Wu, and X.~Lu, ``Denoising prior driven
  deep neural network for image restoration,'' \emph{IEEE transactions on
  pattern analysis and machine intelligence}, vol.~41, no.~10, pp. 2305--2318,
  2018.

\bibitem{guo2019toward}
S.~Guo, Z.~Yan, K.~Zhang, W.~Zuo, and L.~Zhang, ``Toward convolutional blind
  denoising of real photographs,'' in \emph{Proceedings of the IEEE Conference
  on Computer Vision and Pattern Recognition}, 2019, pp. 1712--1722.

\bibitem{chen2019real}
C.~Chen, Z.~Xiong, X.~Tian, Z.-J. Zha, and F.~Wu, ``Real-world image denoising
  with deep boosting,'' \emph{IEEE transactions on pattern analysis and machine
  intelligence}, vol.~42, no.~12, pp. 3071--3087, 2019.

\bibitem{zhang2020residual}
Y.~Zhang, Y.~Tian, Y.~Kong, B.~Zhong, and Y.~Fu, ``Residual dense network for
  image restoration,'' \emph{IEEE Transactions on Pattern Analysis and Machine
  Intelligence}, 2020.

\bibitem{jiang2022deep}
B.~Jiang, Y.~Lu, J.~Wang, G.~Lu, and D.~Zhang, ``Deep image denoising with
  adaptive priors,'' \emph{IEEE Transactions on Circuits and Systems for Video
  Technology}, vol.~32, no.~8, pp. 5124--5136, 2022.

\bibitem{pan2022real}
Y.~Pan, C.~Ren, X.~Wu, J.~Huang, and X.~He, ``Real image denoising via guided
  residual estimation and noise correction,'' \emph{IEEE Transactions on
  Circuits and Systems for Video Technology}, vol.~33, no.~4, pp. 1994--2000,
  2022.

\bibitem{zhou2023deep}
Z.~Zhou, Y.~Chen, and Y.~Zhou, ``Deep dynamic memory augmented attentional
  dictionary learning for image denoising,'' \emph{IEEE Transactions on
  Circuits and Systems for Video Technology}, 2023.

\bibitem{lu2023virtual}
Z.~Lu, Y.~Liu, M.~Jin, X.~Luo, H.~Yue, Z.~Wang, S.~Zuo, Y.~Zeng, J.~Fan,
  Y.~Pang \emph{et~al.}, ``Virtual-scanning light-field microscopy for robust
  snapshot high-resolution volumetric imaging,'' \emph{Nature Methods},
  vol.~20, no.~5, pp. 735--746, 2023.

\bibitem{buades2019cfa}
A.~Buades and J.~Duran, ``Cfa video denoising and demosaicking chain via
  spatio-temporal patch-based filtering,'' \emph{IEEE Transactions on Circuits
  and Systems for Video Technology}, vol.~30, no.~11, pp. 4143--4157, 2019.

\bibitem{sun2023deep}
L.~Sun, Y.~Wang, F.~Wu, X.~Li, W.~Dong, and G.~Shi, ``Deep unfolding network
  for efficient mixed video noise removal,'' \emph{IEEE Transactions on
  Circuits and Systems for Video Technology}, 2023.

\bibitem{matteo2011video}
M.~Matteo, G.~Boracchi, F.~Alessandro, E.~Karen \emph{et~al.}, ``Video
  denoising using separable 4d nonlocal spatiotemporal transforms.'' in
  \emph{Image Processing: Algorithms and Systems IX}.\hskip 1em plus 0.5em
  minus 0.4em\relax SPIE, 2011, pp. 1--11.

\bibitem{chen2016deep}
X.~Chen, L.~Song, and X.~Yang, ``Deep rnns for video denoising,'' in
  \emph{Applications of Digital Image Processing XXXIX}, vol. 9971.\hskip 1em
  plus 0.5em minus 0.4em\relax International Society for Optics and Photonics,
  2016, p. 99711T.

\bibitem{xue2019video}
T.~Xue, B.~Chen, J.~Wu, D.~Wei, and W.~T. Freeman, ``Video enhancement with
  task-oriented flow,'' \emph{International Journal of Computer Vision}, vol.
  127, no.~8, pp. 1106--1125, 2019.

\bibitem{zhang2024binarized}
G.~Zhang, Y.~Zhang, X.~Yuan, and Y.~Fu, ``Binarized low-light raw video
  enhancement,'' in \emph{Proceedings of the IEEE/CVF Conference on Computer
  Vision and Pattern Recognition}, 2024, pp. 25\,753--25\,762.

\bibitem{maggioni2021efficient}
M.~Maggioni, Y.~Huang, C.~Li, S.~Xiao, Z.~Fu, and F.~Song, ``Efficient
  multi-stage video denoising with recurrent spatio-temporal fusion,'' in
  \emph{Proceedings of the IEEE/CVF Conference on Computer Vision and Pattern
  Recognition}, 2021, pp. 3466--3475.

\bibitem{li2022unidirectional}
J.~Li, X.~Wu, Z.~Niu, and W.~Zuo, ``Unidirectional video denoising by mimicking
  backward recurrent modules with look-ahead forward ones,'' in \emph{European
  Conference on Computer Vision}.\hskip 1em plus 0.5em minus 0.4em\relax
  Springer, 2022, pp. 592--609.

\bibitem{tassano2020fastdvdnet}
M.~Tassano, J.~Delon, and T.~Veit, ``Fastdvdnet: Towards real-time deep video
  denoising without flow estimation,'' in \emph{Proceedings of the IEEE/CVF
  Conference on Computer Vision and Pattern Recognition}, 2020, pp. 1354--1363.

\bibitem{qi2022real}
C.~Qi, J.~Chen, X.~Yang, and Q.~Chen, ``Real-time streaming video denoising
  with bidirectional buffers,'' in \emph{Proceedings of the 30th ACM
  International Conference on Multimedia}, 2022, pp. 2758--2766.

\bibitem{li2023simple}
D.~Li, X.~Shi, Y.~Zhang, K.~C. Cheung, S.~See, X.~Wang, H.~Qin, and H.~Li, ``A
  simple baseline for video restoration with grouped spatial-temporal shift,''
  in \emph{Proceedings of the IEEE/CVF Conference on Computer Vision and
  Pattern Recognition}, 2023, pp. 9822--9832.

\bibitem{lehtinen2018noise2noise}
J.~Lehtinen, J.~Munkberg, J.~Hasselgren, S.~Laine, T.~Karras, M.~Aittala, and
  T.~Aila, ``Noise2noise: Learning image restoration without clean data,''
  \emph{arXiv preprint arXiv:1803.04189}, 2018.

\bibitem{krull2019noise2void}
A.~Krull, T.-O. Buchholz, and F.~Jug, ``Noise2void-learning denoising from
  single noisy images,'' in \emph{Proceedings of the IEEE Conference on
  Computer Vision and Pattern Recognition}, 2019, pp. 2129--2137.

\bibitem{pang2021recorrupted}
T.~Pang, H.~Zheng, Y.~Quan, and H.~Ji, ``Recorrupted-to-recorrupted:
  unsupervised deep learning for image denoising,'' in \emph{Proceedings of the
  IEEE/CVF conference on computer vision and pattern recognition}, 2021, pp.
  2043--2052.

\bibitem{ehret2019model}
T.~Ehret, A.~Davy, J.-M. Morel, G.~Facciolo, and P.~Arias, ``Model-blind video
  denoising via frame-to-frame training,'' in \emph{Proceedings of the IEEE
  Conference on Computer Vision and Pattern Recognition}, 2019, pp.
  11\,369--11\,378.

\bibitem{zhang2017beyond}
K.~Zhang, W.~Zuo, Y.~Chen, D.~Meng, and L.~Zhang, ``Beyond a gaussian denoiser:
  Residual learning of deep cnn for image denoising,'' \emph{IEEE Transactions
  on Image Processing}, vol.~26, no.~7, pp. 3142--3155, 2017.

\bibitem{dewil2021self}
V.~Dewil, J.~Anger, A.~Davy, T.~Ehret, G.~Facciolo, and P.~Arias,
  ``Self-supervised training for blind multi-frame video denoising,'' in
  \emph{Proceedings of the IEEE/CVF winter conference on applications of
  computer vision}, 2021, pp. 2724--2734.

\bibitem{sheth2021unsupervised}
D.~Y. Sheth, S.~Mohan, J.~L. Vincent, R.~Manzorro, P.~A. Crozier, M.~M. Khapra,
  E.~P. Simoncelli, and C.~Fernandez-Granda, ``Unsupervised deep video
  denoising,'' in \emph{Proceedings of the IEEE/CVF International Conference on
  Computer Vision}, 2021, pp. 1759--1768.

\bibitem{fu2024temporal}
Z.~Fu, L.~Guo, C.~Wang, Y.~Wang, Z.~Li, and B.~Wen, ``Temporal as a plugin:
  Unsupervised video denoising with pre-trained image denoisers,'' \emph{ECCV},
  2024.

\bibitem{cao2024zero}
C.~Cao, H.~Yue, X.~Liu, and J.~Yang, ``Zero-shot video restoration and
  enhancement using pre-trained image diffusion model,'' \emph{AAAI}, 2025.

\bibitem{xu2019towards}
X.~Xu, Y.~Ma, and W.~Sun, ``Towards real scene super-resolution with raw
  images,'' in \emph{Proceedings of the IEEE Conference on Computer Vision and
  Pattern Recognition}, 2019, pp. 1723--1731.

\bibitem{zhang2019zoom}
X.~Zhang, Q.~Chen, R.~Ng, and V.~Koltun, ``Zoom to learn, learn to zoom,'' in
  \emph{Proceedings of the IEEE Conference on Computer Vision and Pattern
  Recognition}, 2019, pp. 3762--3770.

\bibitem{yue2022real}
H.~Yue, Z.~Zhang, and J.~Yang, ``Real-rawvsr: Real-world raw video
  super-resolution with a benchmark dataset,'' in \emph{European Conference on
  Computer Vision}.\hskip 1em plus 0.5em minus 0.4em\relax Springer, 2022, pp.
  608--624.

\bibitem{ratnasingam2019deep}
S.~Ratnasingam, ``Deep camera: A fully convolutional neural network for image
  signal processing,'' in \emph{Proceedings of the IEEE International
  Conference on Computer Vision Workshops}, 2019, pp. 0--0.

\bibitem{schwartz2018deepisp}
E.~Schwartz, R.~Giryes, and A.~M. Bronstein, ``Deepisp: learning end-to-end
  image processing pipeline,'' \emph{arXiv preprint arXiv:1801.06724}, 2018.

\bibitem{liang2019cameranet}
Z.~Liang, J.~Cai, Z.~Cao, and L.~Zhang, ``Cameranet: A two-stage framework for
  effective camera isp learning,'' \emph{arXiv preprint arXiv:1908.01481},
  2019.

\bibitem{ignatov2020aim}
A.~Ignatov, R.~Timofte, Z.~Zhang, M.~Liu, H.~Wang, W.~Zuo, J.~Zhang, R.~Zhang,
  Z.~Peng, S.~Ren \emph{et~al.}, ``Aim 2020 challenge on learned image signal
  processing pipeline,'' \emph{arXiv preprint arXiv:2011.04994}, 2020.

\bibitem{liang2020raw}
C.-H. Liang, Y.-A. Chen, Y.-C. Liu, and W.~Hsu, ``Raw image deblurring,''
  \emph{IEEE Transactions on Multimedia}, 2020.

\bibitem{yue2022recaptured}
H.~Yue, Y.~Cheng, Y.~Mao, C.~Cao, and J.~Yang, ``Recaptured screen image
  demoir{\'e}ing in raw domain,'' \emph{IEEE Transactions on Multimedia}, 2022.

\bibitem{gharbi2016deep}
M.~Gharbi, G.~Chaurasia, S.~Paris, and F.~Durand, ``Deep joint demosaicking and
  denoising,'' \emph{ACM Transactions on Graphics (TOG)}, vol.~35, no.~6, p.
  191, 2016.

\bibitem{chen2019learning}
C.~Chen, Q.~Chen, M.~N. Do, and V.~Koltun, ``Learning to see in the dark,'' in
  \emph{Proceedings of the IEEE Conference on Computer Vision and Pattern
  Recognition}, 2018.

\bibitem{liu2019learning}
J.~Liu, C.-H. Wu, Y.~Wang, Q.~Xu, Y.~Zhou, H.~Huang, C.~Wang, S.~Cai, Y.~Ding,
  H.~Fan \emph{et~al.}, ``Learning raw image denoising with bayer pattern
  unification and bayer preserving augmentation,'' in \emph{Proceedings of the
  IEEE Conference on Computer Vision and Pattern Recognition Workshops}, 2019,
  pp. 0--0.

\bibitem{abdelhamed2019ntire}
A.~Abdelhamed, R.~Timofte, and M.~S. Brown, ``Ntire 2019 challenge on real
  image denoising: Methods and results,'' in \emph{Proceedings of the IEEE
  Conference on Computer Vision and Pattern Recognition Workshops}, 2019, pp.
  0--0.

\bibitem{abdelhamed2020ntire}
A.~Abdelhamed, M.~Afifi, R.~Timofte, and M.~S. Brown, ``Ntire 2020 challenge on
  real image denoising: Dataset, methods and results,'' in \emph{Proceedings of
  the IEEE/CVF Conference on Computer Vision and Pattern Recognition
  Workshops}, 2020, pp. 496--497.

\bibitem{anaya2014renoir}
J.~Anaya and A.~Barbu, ``Renoir-a dataset for real low-light image noise
  reduction,'' \emph{arXiv preprint arXiv:1409.8230}, 2014.

\bibitem{abdelhamed2018high}
A.~Abdelhamed, S.~Lin, and M.~S. Brown, ``A high-quality denoising dataset for
  smartphone cameras,'' in \emph{Proceedings of the IEEE Conference on Computer
  Vision and Pattern Recognition}, 2018, pp. 1692--1700.

\bibitem{plotz2017benchmarking}
T.~Plotz and S.~Roth, ``Benchmarking denoising algorithms with real
  photographs,'' in \emph{Proceedings of the IEEE Conference on Computer Vision
  and Pattern Recognition}, 2017, pp. 1586--1595.

\bibitem{brooks2018unprocessing}
T.~Brooks, B.~Mildenhall, T.~Xue, J.~Chen, D.~Sharlet, and J.~T. Barron,
  ``Unprocessing images for learned raw denoising,'' \emph{CVPR}, 2019.

\bibitem{zamir2020cycleisp}
S.~W. Zamir, A.~Arora, S.~Khan, M.~Hayat, F.~S. Khan, M.-H. Yang, and L.~Shao,
  ``Cycleisp: Real image restoration via improved data synthesis,'' in
  \emph{Proceedings of the IEEE/CVF Conference on Computer Vision and Pattern
  Recognition}, 2020, pp. 2696--2705.

\bibitem{wang2020practical}
Y.~Wang, H.~Huang, Q.~Xu, J.~Liu, Y.~Liu, and J.~Wang, ``Practical deep raw
  image denoising on mobile devices,'' in \emph{European Conference on Computer
  Vision}.\hskip 1em plus 0.5em minus 0.4em\relax Springer, 2020, pp. 1--16.

\bibitem{chen2019seeing}
C.~Chen, Q.~Chen, M.~N. Do, and V.~Koltun, ``Seeing motion in the dark,'' in
  \emph{Proceedings of the IEEE International Conference on Computer Vision},
  2019.

\bibitem{nam2016holistic}
S.~Nam, Y.~Hwang, Y.~Matsushita, and S.~Joo~Kim, ``A holistic approach to
  cross-channel image noise modeling and its application to image denoising,''
  in \emph{Proceedings of the IEEE Conference on Computer Vision and Pattern
  Recognition}, 2016, pp. 1683--1691.

\bibitem{yue2019high}
H.~Yue, J.~Liu, J.~Yang, T.~Nguyen, and F.~Wu, ``High iso jpeg image denoising
  by deep fusion of collaborative and convolutional filtering,'' \emph{IEEE
  Transactions on Image Processing}, 2019.

\bibitem{xu2018real}
J.~Xu, H.~Li, Z.~Liang, D.~Zhang, and L.~Zhang, ``Real-world noisy image
  denoising: A new benchmark,'' \emph{arXiv preprint arXiv:1804.02603}, 2018.

\bibitem{xu2022pvdd}
X.~Xu, Y.~Yu, N.~Jiang, J.~Lu, B.~Yu, and J.~Jia, ``Pvdd: A practical video
  denoising dataset with real-world dynamic scenes,'' \emph{arXiv preprint
  arXiv:2207.01356}, 2022.

\bibitem{jiang2019learning}
H.~Jiang and Y.~Zheng, ``Learning to see moving objects in the dark,'' in
  \emph{Proceedings of the IEEE/CVF international conference on computer
  vision}, 2019, pp. 7324--7333.

\bibitem{peng2019learned}
Y.~Peng, Q.~Sun, X.~Dun, G.~Wetzstein, W.~Heidrich, and F.~Heide, ``Learned
  large field-of-view imaging with thin-plate optics.'' \emph{ACM Trans.
  Graph.}, vol.~38, no.~6, pp. 219--1, 2019.

\bibitem{perazzi2016benchmark}
F.~Perazzi, J.~Pont-Tuset, B.~McWilliams, L.~Van~Gool, M.~Gross, and
  A.~Sorkine-Hornung, ``A benchmark dataset and evaluation methodology for
  video object segmentation,'' in \emph{Proceedings of the IEEE conference on
  computer vision and pattern recognition}, 2016, pp. 724--732.

\bibitem{mercat2020uvg}
A.~Mercat, M.~Viitanen, and J.~Vanne, ``Uvg dataset: 50/120fps 4k sequences for
  video codec analysis and development,'' in \emph{Proceedings of the 11th ACM
  Multimedia Systems Conference}, 2020, pp. 297--302.

\bibitem{su2017deep}
S.~Su, M.~Delbracio, J.~Wang, G.~Sapiro, W.~Heidrich, and O.~Wang, ``Deep video
  deblurring for hand-held cameras,'' in \emph{Proceedings of the IEEE
  conference on computer vision and pattern recognition}, 2017, pp. 1279--1288.

\bibitem{thongkamwitoon2015image}
T.~Thongkamwitoon, H.~Muammar, and P.-L. Dragotti, ``An image recapture
  detection algorithm based on learning dictionaries of edge profiles,''
  \emph{IEEE Transactions on Information Forensics and Security}, vol.~10,
  no.~5, pp. 953--968, 2015.

\bibitem{wei2020physics}
K.~Wei, Y.~Fu, J.~Yang, and H.~Huang, ``A physics-based noise formation model
  for extreme low-light raw denoising,'' in \emph{Proceedings of the IEEE/CVF
  Conference on Computer Vision and Pattern Recognition}, 2020, pp. 2758--2767.

\bibitem{zhang2021rethinking}
Y.~Zhang, H.~Qin, X.~Wang, and H.~Li, ``Rethinking noise synthesis and modeling
  in raw denoising,'' in \emph{Proceedings of the IEEE/CVF International
  Conference on Computer Vision}, 2021, pp. 4593--4601.

\bibitem{feng2022learnability}
H.~Feng, L.~Wang, Y.~Wang, and H.~Huang, ``Learnability enhancement for
  low-light raw denoising: Where paired real data meets noise modeling,'' in
  \emph{Proceedings of the 30th ACM International Conference on Multimedia},
  2022, pp. 1436--1444.

\bibitem{weinzaepfel2013deepflow}
P.~Weinzaepfel, J.~Revaud, Z.~Harchaoui, and C.~Schmid, ``Deepflow: Large
  displacement optical flow with deep matching,'' in \emph{Proceedings of the
  IEEE international conference on computer vision}, 2013, pp. 1385--1392.

\bibitem{chen2021crossvit}
C.-F.~R. Chen, Q.~Fan, and R.~Panda, ``Crossvit: Cross-attention multi-scale
  vision transformer for image classification,'' in \emph{Proceedings of the
  IEEE/CVF international conference on computer vision}, 2021, pp. 357--366.

\bibitem{chen2022mobile}
Y.~Chen, X.~Dai, D.~Chen, M.~Liu, X.~Dong, L.~Yuan, and Z.~Liu,
  ``Mobile-former: Bridging mobilenet and transformer,'' in \emph{Proceedings
  of the IEEE/CVF Conference on Computer Vision and Pattern Recognition}, 2022,
  pp. 5270--5279.

\bibitem{zhang2022efficient}
X.~Zhang, H.~Zeng, S.~Guo, and L.~Zhang, ``Efficient long-range attention
  network for image super-resolution,'' in \emph{Computer Vision--ECCV 2022:
  17th European Conference, Tel Aviv, Israel, October 23--27, 2022,
  Proceedings, Part XVII}.\hskip 1em plus 0.5em minus 0.4em\relax Springer,
  2022, pp. 649--667.

\bibitem{ding2021diverse}
X.~Ding, X.~Zhang, J.~Han, and G.~Ding, ``Diverse branch block: Building a
  convolution as an inception-like unit,'' in \emph{Proceedings of the IEEE/CVF
  Conference on Computer Vision and Pattern Recognition}, 2021, pp.
  10\,886--10\,895.

\bibitem{paliwal2021multi}
A.~Paliwal, L.~Zeng, and N.~K. Kalantari, ``Multi-stage raw video denoising
  with adversarial loss and gradient mask,'' in \emph{2021 IEEE International
  Conference on Computational Photography (ICCP)}.\hskip 1em plus 0.5em minus
  0.4em\relax IEEE, 2021, pp. 1--10.

\bibitem{ostrowski2022bp}
P.~K. Ostrowski, E.~Katsaros, D.~Wesierski, and A.~Jezierska, ``Bp-evd: Forward
  block-output propagation for efficient video denoising,'' \emph{IEEE
  Transactions on Image Processing}, vol.~31, pp. 3809--3824, 2022.

\bibitem{wang2019edvr}
X.~Wang, K.~C. Chan, K.~Yu, C.~Dong, and C.~Change~Loy, ``Edvr: Video
  restoration with enhanced deformable convolutional networks,'' in
  \emph{Proceedings of the IEEE Conference on Computer Vision and Pattern
  Recognition Workshops}, 2019, pp. 0--0.

\bibitem{fang2022cross}
C.~Fang, D.~Zhang, L.~Wang, Y.~Zhang, L.~Cheng, and J.~Han, ``Cross-modality
  high-frequency transformer for mr image super-resolution,'' in
  \emph{Proceedings of the 30th ACM International Conference on Multimedia},
  2022, pp. 1584--1592.

\bibitem{cheng2024progressive}
D.~Cheng, Y.~Li, D.~Zhang, N.~Wang, J.~Sun, and X.~Gao, ``Progressive negative
  enhancing contrastive learning for image dehazing and beyond,'' \emph{IEEE
  Transactions on Multimedia}, 2024.

\bibitem{cheng2024continual}
D.~Cheng, Y.~Ji, D.~Gong, Y.~Li, N.~Wang, J.~Han, and D.~Zhang, ``Continual
  all-in-one adverse weather removal with knowledge replay on a unified network
  structure,'' \emph{IEEE Transactions on Multimedia}, 2024.

\end{thebibliography}
%
%

%




\end{document}